\documentclass[letterpaper]{article}

\usepackage{style}

\usepackage[hyphens]{url}
\usepackage{graphicx}
\usepackage[numbers,sort&compress]{natbib}
\usepackage{caption}
\usepackage{booktabs}
\usepackage[table]{xcolor}
\usepackage{amsmath}
\usepackage{amssymb}
\usepackage{amsfonts}
\usepackage{multirow}
\usepackage[algo2e,ruled,vlined,linesnumbered]{algorithm2e}

\usepackage[colorlinks=true,linkcolor=blue,citecolor=blue,urlcolor=blue,filecolor=blue]{hyperref}
\hypersetup{pdfstartview={}}

\newcommand{\sectionref}[1]{Section~\ref{#1}}
\newcommand{\acks}[1]{\section*{Acknowledgments}#1}
\newcommand{\tableref}[1]{Table~\ref{#1}}
\newcommand{\figureref}[1]{Figure~\ref{#1}}
\newcommand{\equationref}[1]{Eq.~\ref{#1}}
\newcommand{\appendixref}[1]{Appendix~\ref{#1}}
\newcommand{\algorithmref}[1]{Algorithm~\ref{#1}}

\makeatletter
\newcommand{\floatconts}[3]{%
  \centering
  \def\fc@table{table}%
  \edef\fc@type{\@captype}%
  \ifx\fc@type\fc@table
    #2\label{#1}\par\vspace{2pt}#3%
  \else
    #3\par#2\label{#1}%
  \fi
}
\makeatother

\makeatletter
\newcommand{\appendixheadings}{%
  \expandafter\def\csname secprefix@section\endcsname{}%
  \renewcommand\@seccntformat[1]{%
    \ifcsname secprefix@##1\endcsname Appendix~\fi
    \csname the##1\endcsname.\quad}}
\makeatother

\title{Selective Linkage Precision for Released Medical Vision-Language Models}
\author{
    Soroosh Tayebi Arasteh\corresponding\textsuperscript{\rm 1},
    Mahshad Lotfinia\textsuperscript{\rm 2},
    Sven Nebelung\textsuperscript{\rm 1},
    Daniel Truhn\textsuperscript{\rm 1}
}
\affiliations{
    \textsuperscript{\rm 1}RWTH Aachen University, Aachen, Germany\\
    \textsuperscript{\rm 2}Friedrich-Alexander-Universit\"at Erlangen-N\"urnberg, Erlangen, Germany\\
    soroosh.arasteh@rwth-aachen.de
}

\begin{document}

\maketitle
\thispagestyle{plain}

\begin{abstract}
Hospitals may release de-identified chest radiographs while restricting access to their reports. An attacker who already has reports from the same population and their metadata can use a vision-language model (VLM) to match images to reports. We introduce \textbf{selective linkage precision}, which measures correct matches among the highest-confidence queries after filtering candidates by the available attributes. Across 8 public VLMs and 3 chest radiograph datasets, the maximum whole-pool Recall@1 is 0.40\%. However, this metric averages over every query without attribute filtering. On ReXGradient-160K, MedSigLIP-448 reaches 67.68\% precision at a budget of 100 after filtering by sex, age band, and projection, with a median of 327 candidate reports. On the same eligible query set, an attribute-only policy reaches 24.99\% and the model alone reaches 1.98\%. Adding acquisition year, available in only 1 of the 3 datasets, yields 99.05\% precision and a median of 24 candidates. Uniform guessing on the observed selected queries has an expected precision of 19.91\%. Expanding the primary-setting collection to all 123{,}100 ReXGradient studies lowers precision to 37.23\%. On BioViL-T, the tested head finetuning procedures do not lower linkage below the public checkpoint. Selective linkage precision requires inference without retraining and could help hospitals evaluate which attributes to publish.
\end{abstract}

\section{Introduction}
\label{sec:intro}

\begin{figure*}[t]
\floatconts
  {fig:overview}
  {\caption{MedSigLIP-448 on ReXGradient-160K at $k=100$. Bars show bootstrap precision. Gray rules mark uniform guessing on the observed selected queries.}}
  {\includegraphics[width=6.5in]{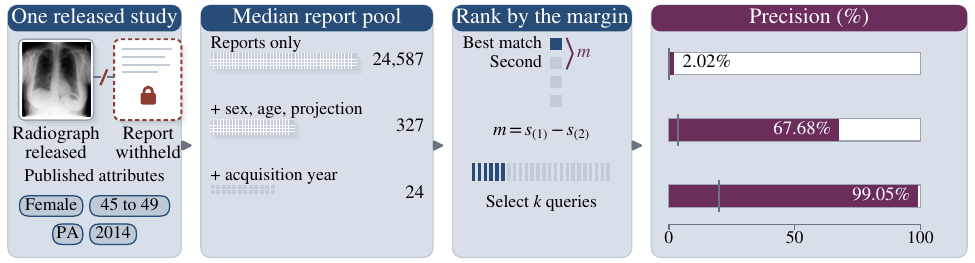}}
\end{figure*}

A hospital may release de-identified chest radiographs while controlling access to the accompanying narrative reports. A researcher who separately holds reports from the same population could try to reconstruct the image-report mapping. If metadata are associated with both collections, the researcher can first retain reports with the image's published sex, age band, and projection. A public vision-language model (VLM) can then rank those candidate reports by image-text similarity. The researcher can also select images whose best match scores well above the runner-up (\figureref{fig:overview}).

Image-to-text retrieval commonly measures how often a model ranks the true report first in a fixed candidate collection \citep{zhang2022convirt,huang2021gloria,Bannur2023Biovilt}. This average scores every query without filtering by the attributes published beside the image. It can obscure an attack that uses these attributes and selects only images with confident matches. We evaluate the precision of that selective attack.

This attack assumes access to an existing report collection and to the report-side attributes used for filtering. We measure correct links within that collection. Obtaining restricted reports, recovering patient names, and extracting training reports from model parameters are separate attacks that this evaluation does not measure.

An attacker can abstain when confidence is low and rank queries without knowing their true pairings. The attributes used for filtering are the published quasi-identifiers that motivated $k$-anonymity \citep{Sweeney2002Kanonymity}. We evaluate selection and attribute filtering together using \textbf{selective linkage precision}, defined as the fraction of correct matches among the $k$ queries with the largest margins between their best and second-best scores. Its uniform-guess reference averages $1/N_i$ over those selected queries, where $N_i$ is the number of candidates for query $i$. Computing the attack requires no retraining or true pairings. Measuring its precision requires ground truth.

Our primary setting uses sex, age band, and projection, which all 3 datasets publish. We additionally evaluate acquisition year on ReXGradient-160K, the only dataset that supplies it. The year-conditioned result describes a more restrictive candidate pool and does not estimate a typical linkage rate.

We make 4 contributions. \textbf{(i)} We define selective linkage precision and evaluate it on 3 datasets and 8 public VLMs. A matched comparison separates model-assisted linkage from an attribute-only policy that selects the smallest candidate pools. \textbf{(ii)} We measure precision with missing correct reports, erroneous attributes, and larger candidate collections. \textbf{(iii)} We estimate associations between report characteristics and recovery after adjustment for other characteristics and candidate-pool size. \textbf{(iv)} We compare linkage and utility for 4 release variants, a matched non-private control, and 7 report transformations. We also examine the scope of differential privacy (DP) accounting for the head training \citep{dwork2014algorithmic}.

\section{Related Work}
\label{sec:related}

\paragraph{Re-Identification and De-Identification of Clinical Data.} The uniqueness of combinations such as postal code, sex, and date of birth motivated $k$-anonymity \citep{Sweeney2002Kanonymity}. \citet{Narayanan2008Deanonymization} used auxiliary data to de-anonymize a released ratings dataset. Networks can also use the biometric information in chest radiographs to verify whether images taken years apart belong to the same patient \citep{Packhauser2022Reidentification}. In pathological speech, speaker anonymization addresses privacy while retaining diagnostic information \citep{TayebiArasteh2024Anonymization}. These studies examine identification within a modality. Our evaluation concerns linkage between images and narrative reports.

\paragraph{Medical Vision-Language Models and Cross-Modal Retrieval.} Contrastive image-text pretraining on paired radiographs and reports \citep{zhang2022convirt,huang2021gloria,Bannur2023Biovilt} builds a shared space in which an image and its report are close. Image-to-text retrieval is the standard evaluation of that alignment. Prior work treats image-to-text retrieval as a capability metric and averages it over a uniform candidate pool. Representational convergence across such models has also been measured \citep{TayebiArasteh2026Convergence}. We instead use the same similarity scores as an adversary's decision rule, at a budget that the adversary chooses and over the pool that remains after conditioning on their side information.

\paragraph{Privacy Attacks and Defenses for Clinical Models.} Membership inference attacks showed that public models can reveal information about their training data \citep{Shokri2017Membership}. Differentially private training is a standard defense \citep{DPSGD}, and prior work has quantified its utility cost in medical imaging \citep{22tayebi2024preserving,13TayebiDomainTransfer,mohammadi2026differential} and the dependence of that cost on the pretraining domain \citep{TayebiArasteh2026Pretraining}. Our linkage task concerns matching available records, including records outside the protected finetuning data. We evaluate how head finetuning affects this attack at both whole-pool and selective operating points.

\section{Measuring Linkage at an Adversarial Operating Point}
\label{sec:method}

\paragraph{Threat Model.}\label{sec:threat} The attacker has released, de-identified images and seeks their corresponding reports in a separately obtained collection from the same population. The report collection is an input to the attack. For example, a researcher with authorized access to reports could obtain an image release of overlapping provenance without its image-report mapping. We do not measure how often this access pattern occurs.

We evaluate 3 settings. Report-only access permits searching the entire collection. Access to structured report metadata additionally permits filtering by sex, age band, and projection, which defines the primary conjunction. The strongest setting adds acquisition year. Filtering requires the attributes to be associated independently with each candidate report, for example through a report-side metadata table or accession records. Keyword mentions in report text provide only partial information about these attributes.

The attacker selects queries using model scores and can abstain on the remaining images. Ground-truth pairings are used only to evaluate the selected matches.

\paragraph{Selective Linkage Precision.}\label{sec:instrument} A vision-language model maps images and reports into a shared embedding space. We score each image-report pair by cosine similarity. Let $\mathcal{R}$ be a collection of distinct report studies, $A$ the available attributes, and $a(i)$ and $a(r)$ their values for query image $i$ and report $r$. The candidate set is $\mathcal{C}_i(A)=\{r\in\mathcal{R}:a(r)=a(i)\}$. With no attributes, $\mathcal{C}_i=\mathcal{R}$. Each study is 1 candidate even when it contains several images.

We rank queries with at least 2 candidates. For each of the $M$ eligible queries, let $N_i=|\mathcal{C}_i|$ and let $s_{(1)}(i)$ and $s_{(2)}(i)$ be the largest and second-largest candidate scores. The \emph{margin}
\begin{equation}
m_i=s_{(1)}(i)-s_{(2)}(i)
\label{eq:margin}
\end{equation}
measures how far the best candidate score exceeds the runner-up. The attacker can compute this margin without the true pairing.

Let $r_i$ be the true report and $c_i$ indicate whether its score is maximal. We count a true-report tie at the maximum as correct, giving an optimistic estimate of attack success. An appendix sensitivity analysis excludes queries where a different report is selected within $10^{-6}$ of the true report's score. For a budget $1\leq k\leq M$, let $\mathcal{S}_k$ contain the $k$ largest-margin queries. The \emph{selective linkage precision} is
\begin{equation}
P(k)=\frac{1}{k}\sum_{i\in\mathcal{S}_k}c_i.
\label{eq:precision}
\end{equation}
At $k=M$, this fraction equals Recall@1. We evaluate budgets of 10, 100, and 1{,}000. Singleton candidate sets are counted separately because they do not permit a margin.

\paragraph{The Reference Level.}\label{sec:chance} Uniform guessing is correct with probability $1/N_i$ on query $i$ when its true report is present. For the selected queries, the expected precision and fold enrichment are
\begin{equation}
\bar p(k)=\frac{1}{k}\sum_{i\in\mathcal{S}_k}\frac{1}{N_i},
\qquad \phi(k)=\frac{P(k)}{\bar p(k)}.
\label{eq:fold}
\end{equation}
This expectation is conditional on the selected queries. Using the entire pool's average or the reciprocal of a median pool size gives a different reference. We report pool-wide folds at Recall@1 separately from selected-query expectations. Reported median pool sizes summarize all eligible queries, not only the selected subset. When a true report is absent, uniform guessing cannot recover it. We evaluate that experiment by precision and the fraction of selected queries with absent reports.

\paragraph{Conditioning on the Published Attributes.}\label{sec:conditioning} We evaluate each available attribute separately and then the primary and year-augmented conjunctions. Candidate filtering precedes score ranking. Studies with several projections match a query when the query's projection is among that study's projections. We use the same attribute mappings for queries and candidate studies.

\paragraph{Isolating the Model from the Attributes.}\label{sec:subtractionmethod} At the primary conjunction, we compare 4 policies on their shared eligible queries. The first matches the attributes and breaks candidate ties with deterministic random scores. The second uses the same candidate predictions but selects queries with the smallest eligible pools, maximizing expected success under uniform guessing. The third ranks the whole collection with the model. The fourth ranks the attribute-filtered collection with the model. We score each policy on the queries that it selects and on those selected by the fourth policy. These comparisons distinguish target selection from report ranking on fixed targets.

The same-query comparison measures report-ranking performance on query images that the combined policy has already chosen. When policies select queries separately, higher precision can reflect easier cases. The 2 sets of precision estimates do not form an additive decomposition into model and metadata contributions.

\paragraph{The Covariate Models.}\label{sec:mapmethod} We first fit each report covariate separately to the margin using ordinary least squares, $m_i=\beta_0+\beta_j x_j(r_i)+\varepsilon_i$. Covariates are standardized, and standard errors are clustered by patient. Each query image is an observation, including images from the same study. For sign agreement across models, we include models with at least 10 correct matches and a Recall@1 fold of at least 2.0.

We also fit the 9 report covariates jointly. Under conditioning, we add the logarithm of the candidate-pool size as a tenth term. We fit a separate model for the margin and for a binary indicator of a correct link. We report variance inflation factors to assess collinearity. These regressions measure associations, including associations with pool size, without identifying causes of recovery.

We derive these covariates from the true reports for retrospective analysis. The attacker does not receive those reports as known matches or use their covariates to select queries.

\section{Experiments}
\label{sec:experiments}

\subsection{Setup}
\label{sec:setup}

\paragraph{Datasets.} Our MIMIC-CXR pool contains 215{,}186 images \citep{johnson2019mimiccxr}. We fit the release variants on its training split. Our other pools contain 157{,}745 CheXpert Plus images from a second academic center with a different reporting style \citep{irvin2019chexpert,chambon2024chexpertplusaugmentinglarge} and 128{,}553 ReXGradient-160K images from 70 institutions \citep{Zhang2025Rexgradient}. All 3 datasets provide the image-report pairing, which lets us determine whether a candidate match is correct. We keep frontal projections only, split by patient, and verify at build time that the patient sets are disjoint. The candidate collection is a dataset's test split by default, containing 38{,}686, 31{,}631, and 24{,}587 studies, and \sectionref{sec:costs} enlarges it. MIMIC-CXR provides sex, age, and projection, CheXpert Plus provides sex and age and has a single projection category, and ReXGradient-160K adds a calendar year and a site. MIMIC-IV shifts each patient's dates by a fixed offset. MIMIC-CXR's study year contains patient-specific date-shift information that we exclude from the primary setting (\appendixref{apd:datasets}). The \emph{primary conjunction} is sex, 5-year age band, and projection, which all 3 datasets support, and the \emph{tightest conjunction} adds the study year.

\paragraph{Model Panel.} We evaluate 8 public VLMs spanning several training domains and architectures (\tableref{tab:panel}). General-domain models include 2 CLIP sizes \citep{pmlr-v139-radford21a} and SigLIP-2 \citep{Tschannen2025Siglip2}. PubMedCLIP \citep{eslami-etal-2023-pubmedclip} and BiomedCLIP \citep{zhang2024biomedclip} use biomedical pretraining. MedSigLIP \citep{Sellergren2025Medgemma} and BioViL-T \citep{Bannur2023Biovilt} include chest radiographs in pretraining. MONET \citep{Kim2024Monet}, trained on dermatology, is an off-domain medical control. Differences between these models also include patch size, training data, and input resolution. Therefore, the panel does not isolate causal effects of model scale, training domain, or loss function. The baselines are record linkage on published attributes and a TorchXRayVision DenseNet-121 \citep{Cohen2022Torchxrayvision}. The classifier predicts 12 of the 14 report findings and ranks reports by cosine similarity between its predictions and the report labels.

\paragraph{Implementation and Reporting.} Embedding extraction and head training use 1 NVIDIA L40S accelerator per job. Report rewriting and labeling use a locally served language model. Embeddings are $L_2$ normalized, making their inner product equal to cosine similarity. Reported rates are percentages. Linkage uses 1{,}000 patient resamples with query selection repeated within each resample. We report their mean, standard deviation, and 95\% percentile interval where uncertainty is shown. Counts, guessing expectations, and folds describe the observed query sets. We adjust $p$ for the false discovery rate (FDR) with Benjamini-Hochberg \citep{Benjamini1995Fdr} within each analysis family.

A patient draw includes all of that patient's query images, retaining dependence between images within and across studies. Paired policy comparisons use the same patient draws. Models and candidate-report collections stay fixed while queries are selected again in each replicate. The intervals describe variation under resampling of the evaluated patients, conditional on those models and collections. They do not include uncertainty about which reports an attacker can obtain.

\begin{table*}[t]
\floatconts
  {tab:panel}
  {\caption{Precision at $k=1{,}000$ (percent). Full/Cond.: unconditioned/primary conjunction. Fold uses observed Recall@1. Green marks folds $>2$. Boldface marks the column maximum.}}
  {\scriptsize
   \setlength{\tabcolsep}{2.6pt}
   \begin{tabular}{@{}lrrrrrrrrrr@{}}
   \toprule
   & & \multicolumn{3}{c}{MIMIC-CXR} & \multicolumn{3}{c}{CheXpert Plus} & \multicolumn{3}{c}{ReXGradient-160K} \\
   \cmidrule(lr){3-5}\cmidrule(lr){6-8}\cmidrule(lr){9-11}
   Release & Params (M) & Full$\downarrow$ & Cond.$\downarrow$ & Fold$\downarrow$ & Full$\downarrow$ & Cond.$\downarrow$ & Fold$\downarrow$ & Full$\downarrow$ & Cond.$\downarrow$ & Fold$\downarrow$ \\
   \midrule
   \multicolumn{11}{@{}l}{\textbf{General domain}}\\
   CLIP ViT-B/32 & 151.3 & 0.00 & 0.20 & 1.22 & 0.00 & 0.00 & 1.29 & 0.00 & 1.90 & 1.24 \\
   CLIP ViT-L/14 & 427.6 & 0.00 & 0.40 & 1.41 & 0.00 & 0.41 & 1.75 & 0.11 & 3.63 & 1.74 \\
   SigLIP-2 L/16 & 882.3 & 0.00 & 0.87 & 1.69 & 0.00 & 0.10 & 1.23 & 0.10 & 3.01 & 1.76 \\
   \midrule
   \multicolumn{11}{@{}l}{\textbf{Biomedical}}\\
   PubMedCLIP & 151.3 & 0.00 & 0.81 & 1.34 & 0.00 & 0.19 & 1.14 & 0.10 & 3.11 & 1.33 \\
   BiomedCLIP & 195.9 & 0.00 & 3.74 & \cellcolor{green!22}5.71 & 0.00 & 1.31 & \cellcolor{green!22}7.13 & 0.10 & 9.13 & \cellcolor{green!22}4.04 \\
   \midrule
   \multicolumn{11}{@{}l}{\textbf{Medical and chest radiology}}\\
   MedSigLIP-448 & 878.3 & 0.85 & \textbf{14.94} & \cellcolor{green!22}\textbf{18.86} & \textbf{0.36} & \textbf{6.42} & \cellcolor{green!22}\textbf{21.98} & 0.90 & \textbf{28.74} & \cellcolor{green!22}\textbf{12.59} \\
   BioViL-T & 137.6 & 0.20 & 8.96 & \cellcolor{green!22}15.95 & 0.00 & 1.40 & \cellcolor{green!22}9.13 & 0.10 & 10.37 & \cellcolor{green!22}6.46 \\
   \midrule
   \multicolumn{11}{@{}l}{\textbf{Off-domain control}}\\
   MONET & 427.6 & 0.00 & 0.38 & 1.05 & 0.00 & 0.26 & 1.05 & 0.00 & 1.53 & 1.10 \\
   \midrule
   \multicolumn{11}{@{}l}{\textbf{No VLM}}\\
   Published attributes & N/A & \textbf{1.11} & N/A & N/A & 0.30 & N/A & N/A & \textbf{4.37} & N/A & N/A \\
   14-label classifier & 7.0 & 0.00 & 0.41 & 1.46 & 0.10 & 0.38 & 1.32 & 0.00 & 1.91 & 1.35 \\
   \midrule
   Recall@1 chance & N/A & 0.0026 & 0.1802 & 1.0 & 0.0032 & 0.1010 & 1.0 & 0.0041 & 0.4718 & 1.0 \\
   \bottomrule
   \end{tabular}}
\end{table*}

\subsection{Selection and Conditioning}
\label{sec:conditioned}

Over the whole pool, Recall@1 is at or below 0.10\% in 20 of the 24 model-dataset combinations. Its maximum is 0.40\%, for MedSigLIP-448 on ReXGradient-160K. Selecting this model's 100 highest-margin queries yields 2.02\% precision from 24{,}587 candidate studies. At a budget of 1{,}000, the maximum selective linkage precision is 0.90\%. Sixteen of the 24 combinations remain at or below 0.10\% before rounding (\tableref{tab:panel}, Full columns). The unconditioned folds at Recall@1 on MIMIC-CXR are 68.01 for BioViL-T and 113.65 for MedSigLIP-448. These large folds coexist with low absolute precision because unconditioned chance is 1 in 38{,}686.

An adversary can select images and use their published attributes. The primary conjunction uses the 3 attributes that all 3 datasets publish. It reduces the median candidate pool from 38{,}686 studies to 862 on MIMIC-CXR, from 31{,}631 to 1{,}203 on CheXpert Plus, and from 24{,}587 to 327 on ReXGradient-160K.

For MedSigLIP-448 on ReXGradient-160K, whole-pool Recall@1 is 0.40\%, and the 100 highest-margin queries give 2.02\%. Under the primary conjunction, Recall@1 is 5.94\% and selective precision at $k=100$ is 67.68\%, with a median pool of 327 studies across eligible queries. Selection and filtering each increase precision in this comparison. At a budget of 1{,}000, selective precision is 6.42\% on CheXpert Plus, 14.94\% on MIMIC-CXR, and 28.74\% on ReXGradient-160K (\tableref{tab:panel}, Cond. columns). On the 100 observed selected queries, an adversary who guesses uniformly inside each candidate set is correct 3.57\% of the time (\figureref{fig:reversal}a).

ReXGradient-160K is the only dataset of our 3 that publishes acquisition year. Adding it reduces the median pool to 24 studies, and among the 100 queries with the largest margins 99 are linked correctly with MedSigLIP-448 and 91 with BioViL-T. This is the strongest tested metadata setting. A uniform guess on MedSigLIP-448's selected queries is already correct 19.91\% of the time, giving a fold of 4.97. The pool-wide expectation of 6.05\% would instead give 16.36 and would overstate enrichment on those queries. On MIMIC-CXR, the shifted-year information raises the selected-query guessing expectation to 41.27\%. We report this result separately because patient-specific date shifting changes what the year reveals (\appendixref{apd:limits}).

For BioViL-T on ReXGradient-160K, the fold at Recall@1 decreases from 6.46 under the primary conjunction to 2.97 when acquisition year is added. The increase in absolute precision does not imply an increase in enrichment over the corresponding guessing expectation.

\begin{table}[t]
\floatconts
  {tab:matched}
  {\caption{Precision at $k=100$ (percent), MedSigLIP-448 on ReXGradient-160K.}}
  {\scriptsize
   \setlength{\tabcolsep}{2.4pt}
   \begin{tabular}{@{}lrr@{}}
   \toprule
   Adversary & Own queries & Combined queries \\
   \midrule
   Attributes, matching set & 18.66 & 0.00 \\
   Attributes, smallest pool first & 24.99 & 0.00 \\
   Release alone, whole pool & 1.98 & 17.79 \\
   Release and attributes & \textbf{67.68} & \textbf{67.68} \\
   \midrule
   Uniform guess & & 3.57 \\
   \bottomrule
   \end{tabular}}
\end{table}

\subsection{Linkage Added by the Release}
\label{sec:subtraction}

On ReXGradient-160K, attribute matching has a Recall@1 of 0.49\%, compared with 0.40\% for the highest-scoring unconditioned model. To compare their selective performance, \tableref{tab:matched} restricts all policies to the 25{,}728 shared eligible queries at the primary conjunction and a budget of 100. The smallest-pool attribute policy reaches 24.99\% precision, the unconditioned MedSigLIP-448 policy reaches 1.98\%, and their combination reaches 67.68\%. The combined attack exceeds the first 2 by 42.69 and 65.69 percentage points, respectively, with adjusted $p=0.0018$ for each difference. On the queries selected by the combined attack, the attributes alone reach 0.00\% and the unconditioned model reaches 17.79\%. This comparison keeps target selection fixed within each resample.

At Recall@1 under the primary conjunction, general-domain models and the off-domain control have folds from 1.05 to 1.76. The 14-label classifier has folds from 1.32 to 1.46. In comparison, the ranges are 4.04 to 7.13 for BiomedCLIP, 6.46 to 15.95 for BioViL-T, and 12.59 to 21.98 for MedSigLIP-448 (\figureref{fig:reversal}b). These are whole-split enrichments, distinct from the selected-query expectations in \tableref{tab:matched}. The maximum fold differences between the 2 CLIP sizes and between SigLIP-2 and CLIP ViT-L/14 are 0.49 and 0.52, respectively.

On ReXGradient-160K, the classifier's selective precision is 1.91\% at a budget of 1{,}000. Its fold at Recall@1 is 1.35, measured over all eligible queries. These are different operating points. \appendixref{apd:nomodel} describes both computational baselines.

\subsection{The Costs of the Attack}
\label{sec:costs}

The rates above assume a complete collection, correct attributes, and 1 dataset's test split. We relax each for MedSigLIP-448 at the primary conjunction and a budget of 100 (\appendixref{apd:costsdetail}).

Missing-report tests remove the true report from an affected query's candidate set. Attribute-error tests perturb query metadata before filtering, which can exclude the true report. Whole-dataset tests retain test-split query images while adding reports from the training and validation splits.

On ReXGradient-160K, expanding the collection to the whole dataset reduces precision from 67.68\% to 37.23\%. A 5\% per-attribute error rate reduces it to 41.26\%, and removing the correct report for a quarter of queries reduces it to 55.97\%. The whole-dataset reductions on ReXGradient-160K and MIMIC-CXR each have adjusted $p=0.0015$. For a release containing the whole dataset, these larger collections better represent the available candidates. The test-split estimate is 1.8 times the whole-dataset estimate on ReXGradient-160K. At a 20\% attribute error rate, the median pool changes from 327 studies to 298, indicating that the decrease in precision does not require a larger pool.

Adding candidate reports changes query margins and can change the selected images. The observed reduction with larger collections is specific to these experiments. It does not bound linkage in another institution's collection, where report composition and available attributes may differ.

When the correct report is absent for a quarter or half of the ReXGradient-160K queries, 19.89\% or 45.98\% of the selected queries, respectively, have absent reports and produce incorrect links. The margin distinguishes report presence from absence with an area under the receiver operating characteristic curve (AUROC) of 49.86\% to 50.17\% across the 9 model-dataset combinations. Several intervals exclude 50\%, but the estimated departures from chance are small. Report text provides limited attribute information. Keyword patterns detect sex, age, and projection mentions in only 4\%, 5\%, and 13\% of ReXGradient-160K reports, respectively (\appendixref{apd:costsdetail}).

\begin{table*}[t]
\floatconts
  {tab:ladder}
  {\caption{Conditioned linkage and MIMIC-CXR utility. T2I counts correct text-to-image matches among 5{,}000 studies. Other values are percentages. Image denotes the image-side classifier.}}
  {\scriptsize
   \setlength{\tabcolsep}{3.6pt}
   \begin{tabular}{@{}lrrrrrrrrr@{}}
   \toprule
   & \multicolumn{3}{c}{$k=100$$\downarrow$} & \multicolumn{3}{c}{$k=1{,}000$$\downarrow$} & \multicolumn{3}{c}{Utility$\uparrow$} \\
   \cmidrule(lr){2-4}\cmidrule(lr){5-7}\cmidrule(lr){8-10}
   Variant & MIMIC & CheXpert & ReXGrad. & MIMIC & CheXpert & ReXGrad. & Zero-shot & T2I count & Image \\
   \midrule
   Public checkpoint & 22.18 & 2.01 & 24.10 & 8.96 & 1.40 & 10.37 & 68.86 & 73 & 78.77 \\
   Matched non-private control & 62.89 & 24.45 & 52.16 & 28.97 & 10.76 & 27.00 & 69.66 & 116 & 77.49 \\
   DP heads & 46.66 & 13.82 & 42.16 & 18.26 & 5.82 & 18.74 & 69.89 & 77 & 78.41 \\
   DP heads, in-batch loss & 47.92 & 5.93 & 41.15 & 20.05 & 3.63 & 18.66 & 69.12 & 108 & 79.08 \\
   Bypass heads on public data & 81.87 & 34.35 & 68.80 & 45.55 & 18.11 & 31.35 & 69.70 & 306 & 80.21 \\
   Random orthogonal remapping & 0.00 & 0.96 & 7.90 & 0.16 & 0.10 & 1.20 & 48.38 & 2 & 78.80 \\
   \bottomrule
   \end{tabular}}
\end{table*}

\subsection{Model-Side and Report-Side Defenses}
\label{sec:defense}

The defense experiments use BioViL-T as their only base model. At the primary conjunction, 41 of 63 paired contrasts are significant after correction, compared with 3 of 63 without conditioning. \tableref{tab:ladder} compares the public checkpoint, the finetuned heads, and their utility on MIMIC-CXR.

At a budget of 100, the public checkpoint reaches 2.01\%, 22.18\%, and 24.10\% precision on CheXpert Plus, MIMIC-CXR, and ReXGradient-160K. The non-private control reaches 24.45\%, 62.89\%, and 52.16\%, respectively. These increases of 22.44, 40.71, and 28.07 percentage points are significant on all 3 datasets. The private run reaches 13.82\%, 46.66\%, and 42.16\%, remaining above the public checkpoint. Relative to the non-private control, the differences are $-10.63$, $-16.23$, and $-10.00$ percentage points, with adjusted $p=0.0495$, $0.0240$, and $0.0990$. At a budget of 1{,}000, all 3 private-control contrasts are significant. The zero-shot macro AUROCs on MIMIC-CXR are 69.89\% for the private run, 69.66\% for the control, and 68.86\% for the checkpoint.

These comparisons do not isolate a causal effect of clipping and noise. The private run selects the checkpoint with the lowest validation linkage. The non-private control selects the highest. Private training also uses Poisson sampling. The control uses shuffled fixed-size batches. The shared architecture, objective, nominal batch size, optimizer, and training duration provide a partial control for finetuning. None of the tested head finetuning procedures lowered linkage below the public checkpoint.

Refitting bypass heads on public paired data produces 34.35\%, 81.87\%, and 68.80\% precision on CheXpert Plus, MIMIC-CXR, and ReXGradient-160K at a budget of 100. An attacker can fit another alignment above the frozen encoders when public paired data are available. Independent random orthogonal maps reduce precision to 0.96\%, 0.00\%, and 7.90\%. They also reduce zero-shot AUROC to 48.38\%, while the image-side classifier has 78.80\% AUROC, compared with 78.77\% for the public checkpoint (\figureref{fig:frontiers}a). The maps preserve within-modality geometry but disrupt image-text alignment.

The privacy accountant concerns head finetuning conditional on fixed encoders and a fixed validation negative set. It does not cover encoder pretraining or protect the validation records. The in-batch objective does not satisfy the per-example decomposition required by the reported clipping argument. For the decomposable objective, the full-run example-level bound is $\varepsilon=1.99$ at $\delta=6\times10^{-6}$. Applying group privacy with the unrounded accountant value and 124 images per patient gives $\varepsilon'=246.96$ and a $\delta'$ of order $10^{103}$, making this generic patient-level bound vacuous.

The protected unit is a training image. Linking a test image to an available report differs from determining whether a record participated in head training. DP bounds the influence of a training record on the released parameters, without requiring low similarity between a test image and its report. Persistent test-set linkage does not demonstrate a violation of the example-level guarantee.

On ReXGradient-160K, MedSigLIP-448 has selective precision of 28.74\% at $k=1{,}000$ for the original reports. The 6 narrative transformations yield estimates from 23.20\% to 36.37\% (\figureref{fig:frontiers}b). Their intervals overlap the original-report interval, except for retaining only the findings and impression, whose estimate is higher. The 14-label template gives 1.99\%. After applying the sensitivity exclusion of \appendixref{apd:ties}, BioViL-T's Recall@1 under this template is 0.45\%, 0.17\%, and 1.23\% on MIMIC-CXR, CheXpert Plus, and ReXGradient-160K, respectively.

Content evaluation uses the same fixed study sample for every transformation. The same language model labels each original and transformed report. Agreement is computed only over findings labeled in both versions. The labeler treats uncertainty as absence and excludes unmentioned findings. We also measure the fraction of originally asserted findings that remain asserted. Additional content axes are laterality, severity, measurements, devices, prior comparisons, and hedging. For each axis, retention is the fraction of reports with an original mention whose transformed report still mentions that axis (\appendixref{apd:transforms}). Shared errors in the rewriting and labeling model could conceal content loss.

Finding agreement does not establish preservation of clinical content. Across the 3 datasets, agreement after relabeling ranges from 86.24\% to 99.39\%. On ReXGradient-160K, the template has 98.95\% agreement but no detected mentions on any of these 6 axes. The language-model rewrite has 97.74\% agreement. Prior-comparison mentions remain in 11\% and hedging mentions in 59\% of the reports originally stating them. These text-pattern measures detect whether a type of detail is mentioned, without verifying that its value or meaning is unchanged. Clinical adequacy was not assessed by clinicians.

\subsection{Report Covariates and Exposure}
\label{sec:exposure}

We examine associations between report characteristics and linkage using patient-clustered regressions (\appendixref{apd:map}). The unconditioned data contain 12{,}419, 11{,}570, and 17{,}593 patient groups on MIMIC-CXR, CheXpert Plus, and ReXGradient-160K, respectively. The corresponding conditioned sets can exclude singleton queries. On ReXGradient-160K, longer reports and mentions of laterality, devices, comparisons, measurements, and positive findings have positive associations with the margin across the 3 included clinical models.

We fit the 9 covariates jointly with log candidate-pool size as a tenth term (\appendixref{apd:adjusted}). For MedSigLIP-448 on ReXGradient-160K, the fit includes 17{,}592 patient groups, and variance inflation factors range from 1.00 to 1.66. Log pool size has the largest standardized coefficient magnitude on correct-link probability. Its coefficient is $-$3.91 percentage points. Six report characteristics remain significant: laterality, devices, positive findings, sections, prior comparisons, and rare findings. Report length is not significant after adjustment. We do not interpret inference for the placeholder term because only 1 query has a positive value in this dataset.

For the 8 covariates other than the sparse placeholder count, margin-slope signs agree across the 3 clinical models for all 8 on ReXGradient-160K at both operating points. At the primary conjunction, 3 of 8 agree on MIMIC-CXR and 4 of 8 on CheXpert Plus. The broad sign agreement is specific to ReXGradient-160K. Cross-model overlap in recovered queries is also reported descriptively because candidate-pool size remains a confounder (\appendixref{apd:overlap}).

The descriptive exposure contrasts show corresponding differences in recovered reports. On ReXGradient-160K at the primary conjunction, BioViL-T's recovered queries have reports that average 18.52 whitespace tokens longer, mention laterality 27.35 percentage points more often, and contain 0.98 more positive findings than unrecovered queries. These contrasts do not establish effects specific to the model. \appendixref{apd:exposure} also reports the near-chance CLIP ViT-B/32 baseline.

\subsection{Ablation Studies}
\label{sec:ablations}

\paragraph{(1) The Decision Statistic and the Budget.} The top-1 score has lower selective precision than the margin in 8 of 9 model-dataset combinations. On CheXpert Plus, the precision ordering of BiomedCLIP and BioViL-T reverses between the ratio and the margin. The ratio is also undefined on up to 31\% of queries under the label template. We use the margin (\appendixref{apd:statistic2}).

\paragraph{(2) The Pathology-Matched Pool.} Replacing the uniform pool with the 10{,}000 candidates closest in the 14-finding label space yields Recall@1 from 0.01\% to 0.60\%. This control reduces variation in those coarse labels without eliminating all diagnostic or report-content differences.

\paragraph{(3) General-Domain and Label Baselines.} On ReXGradient-160K with acquisition year added, CLIP ViT-B/32 correctly matches 31 of 100 selected queries and the 14-label classifier matches 24, with a median pool of 24 studies. Their folds at Recall@1 are 1.06 and 1.25, compared with 2.97 for BioViL-T and 4.13 for MedSigLIP-448. Small conditioned pools permit correct matches even for baselines with little enrichment over guessing.

\paragraph{(4) Ties and Uniquely Identified Queries.} Counting true-report ties as correct gives an optimistic attack estimate. A sensitivity analysis excludes queries where another study is selected within $10^{-6}$ of the true score. Up to 1{,}964 queries meet this rule under the label template. Singleton pools are counted separately (\appendixref{apd:ties}).

\paragraph{(5) Agreement of the Release Rankings.} Under the primary conjunction, Kendall's $\tau$ between Recall@1 and selective precision rankings ranges from 0.697 to 0.939 at a budget of 1{,}000 and from 0.574 to 0.931 at a budget of 100. The rankings are positively associated but contain inverted pairs (\appendixref{apd:tau}).

\paragraph{(6) Several Images of One Study.} We average candidate scores across images of the same study and projection, then compare this attack with the largest-margin single image from each group. The datasets contain 572 to 4{,}239 eligible groups with several images. None of the 21 paired differences is significant after FDR correction (\appendixref{apd:studylevel}).

\section{Discussion and Conclusion}
\label{sec:conclusion}

Whole-pool Recall@1 has a maximum of 0.40\% across the evaluated models and datasets. In comparison, MedSigLIP-448 reaches 67.68\% precision at a budget of 100 on ReXGradient-160K after filtering by sex, age band, and projection. This increase combines query selection with candidate filtering. It assumes an existing report collection and independently associated report metadata. The primary matched comparison shows higher precision for their combined use than for the model-only or attribute-only policies.

In the tested conditions, enlarging the collection, adding attribute errors, or removing correct reports lowers precision. Head finetuning on BioViL-T raises linkage above the public checkpoint. The private run has lower precision estimates than the non-private control, but different checkpoint selection and batch sampling prevent causal attribution to clipping and noise. These experiments do not establish equivalence in clinical utility.

Pretraining overlap limits the interpretation of the held-out evaluation. Patients are held out from head finetuning. Their exclusion from encoder pretraining is not established. BioViL-T and MedSigLIP include MIMIC-CXR in their stated pretraining sources \citep{Bannur2023Biovilt,Sellergren2025Medgemma}. Public model descriptions do not permit an exact record-level overlap audit for every checkpoint. Correct linkage can reflect generalization, pretraining overlap, or a combination of these mechanisms. Our evaluation does not distinguish these explanations.

The evaluation is limited to English-language frontal chest radiographs and BioViL-T defense experiments (\appendixref{apd:limits}). We performed no reader study or clinician assessment of transformed reports.

For a hospital evaluating a release, the candidate collection should approximate the reports that an attacker could obtain. The hospital can compare proposed attribute combinations, count singleton pools, and evaluate selective precision beside whole-pool Recall@1. A large margin ranks a query highly but does not establish that its correct report is present. The hospital also needs to test incomplete report collections. A release decision requires assessing the utility of the images and reports after any proposed defense (\appendixref{apd:practice}).

Selective linkage precision needs inference without retraining. A hospital with the true pairings and report metadata can use it to compare planned releases before deciding which attributes to publish.

\acks{STA is supported by the German Research Foundation (DFG; 585806541), the German Federal Ministry of Research, Technology and Space (ARISTOTLE, 01ZU2602B), and the Excellence Strategy of the German Federal Government, the L\"ander, and RWTH ERS (START\_526-26).}
\label{end:body}

\medskip

\paragraph*{Data and Code Availability} MIMIC-CXR is available to credentialed PhysioNet users, CheXpert Plus from Stanford AIMI under a research use agreement, and ReXGradient-160K under its non-commercial data access and use agreement. All models are open weight. The code is available at \url{https://github.com/tayebiarasteh/crossmodal}.

\paragraph*{Institutional Review Board (IRB)} This work is a secondary analysis of existing de-identified public datasets. No new data were collected and no patient was contacted. No IRB review was required.

\paragraph*{Generative AI Disclosure} During the preparation of this work, the authors used Claude Opus 5, GPT-5.6 Sol, and GPT-6 Astra for language polishing. After using these tools, the authors reviewed and edited the content as needed.

\bibliography{refs}

\clearpage
\twocolumn[{\centering\Large\bfseries Supplementary Information\par\vspace{1.2em}}]
\appendix
\appendixheadings
\setcounter{figure}{0}\renewcommand{\thefigure}{S\arabic{figure}}
\setcounter{table}{0}\renewcommand{\thetable}{S\arabic{table}}
\setcounter{equation}{0}\renewcommand{\theequation}{S\arabic{equation}}
\setcounter{algocf}{0}\renewcommand{\thealgocf}{S\arabic{algocf}}

\section{Datasets, Models, and Adversaries}
\label{apd:assets}

The appendix is grouped by topic. Datasets and adversaries are described in \appendixref{apd:assets}, linkage estimation and report transformations in \appendixref{apd:measurement}, and head training in \appendixref{apd:training}. Extended results and robustness analyses appear in \appendixref{apd:extended} and \appendixref{apd:further}. Limitations and hospital evaluation are discussed in \appendixref{apd:limits} and \appendixref{apd:practice}.

\subsection{Dataset Construction and the Splits}
\label{apd:datasets}

\paragraph{Sources and the Label Space.} MIMIC-CXR provides 1 report file per study, from which we parse the sections. CheXpert Plus provides separate section fields for the findings and the impression. ReXGradient-160K provides its reports inline. We use 14 findings: atelectasis, cardiomegaly, consolidation, edema, enlarged cardiomediastinum, fracture, lung lesion, lung opacity, no finding, pleural effusion, pleural other, pneumonia, pneumothorax, and support devices. For MIMIC-CXR and CheXpert Plus, present findings count as positive, absent or unmentioned findings count as negative, and uncertain findings are excluded per finding. ReXGradient-160K has no supplied finding labels. Its language-model labels count present findings as positive, merge absent and uncertain findings as negative, and exclude unmentioned findings, as specified by the quoted prompt in \appendixref{apd:transforms}.

\paragraph{The Study as the Unit of the Candidate Pool.} We use 1 candidate per study, retaining the first report encountered for that study. MIMIC-CXR and ReXGradient-160K provide study identifiers. For CheXpert Plus, we derive them from the patient and study fields encoded in each image path. Its 2{,}704 groups with several frontal images contain different full report texts, and 792 also differ in findings or impression text. The retained report represents the group. Linkage to this representative does not establish recovery of every image-specific report version.

\paragraph{Preexisting Cohort Filters.} The supplied MIMIC-CXR source list excludes images with uncertain labels for atelectasis, consolidation, enlarged cardiomediastinum, pleural effusion, or pneumothorax. The CheXpert source list excludes images with uncertain labels for cardiomegaly, lung opacity, lung lesion, edema, or pneumonia. These image exclusions precede our frontal-image selection and patient split. Uncertainty in the remaining findings is handled per finding as described above. Therefore, the evaluated pools differ from the complete dataset releases.

PA and AP denote posteroanterior and anteroposterior projections, respectively.

\paragraph{Source Resolution.} Every model receives the same preprocessed source images at $224 \times 224$ pixels, resized to its required input dimensions. This choice controls source resolution across models and datasets. The global random seed is 42.

\begin{table}[t]
\floatconts
  {tab:datasets}
  {\caption{The 3 datasets, their splits, and the attributes that they publish.}}
  {\scriptsize
   \setlength{\tabcolsep}{2pt}
   \begin{tabular}{@{}lrrr@{}}
   \toprule
   & MIMIC & CheXpert & ReXGrad. \\
   \midrule
   Images in the pool & 215{,}186 & 157{,}745 & 128{,}553 \\
   Patients & 62{,}094 & 57{,}848 & 87{,}965 \\
   Studies & 192{,}701 & 155{,}012 & 123{,}100 \\
   \midrule
   Test queries & 43{,}231 & 32{,}210 & 25{,}732 \\
   Distinct studies & 38{,}686 & 31{,}631 & 24{,}587 \\
   Test patients & 12{,}419 & 11{,}570 & 17{,}593 \\
   \midrule
   Sex & yes & yes & yes \\
   Age band & yes & yes & yes \\
   Projection & PA, AP & frontal & PA, AP \\
   Study year & shifted & absent & 2004--2024 \\
   Site & N/A & N/A & 70 \\
   \bottomrule
   \end{tabular}}
\end{table}

\tableref{tab:datasets} lists the final pool sizes and the attributes available for conditioning. Studies whose report is missing or shorter than 20 characters are dropped at build time.

We construct patient-disjoint splits with seed 20260222. We reserve 20\% of patients for testing and 10\% of the remaining patients for validation, yielding 72\%, 8\%, and 20\% for training, validation, and testing. The default candidate collection contains only reports from that dataset's test split. The larger-collection experiments additionally include the other test splits or the training and validation reports of the same dataset (\appendixref{apd:costsdetail}). ReXGradient-160K includes 79 institutions in its published description and 70 in our frontal-image pool.

MIMIC-CXR's shifted study years take 99 values from 2110 to 2208. The date offset is fixed per patient. Longitudinal studies can have different shifted years. These years do not represent acquisition calendar years.

\subsection{The Model Panel}
\label{apd:panel}

\paragraph{Model Loading and Checkpoint Details.} Six models use the Transformers image-text interface. BiomedCLIP uses OpenCLIP, and BioViL-T uses its dedicated image inference implementation and text encoder. BioViL-T resizes the shorter image edge to 512 pixels and center-crops to $448 \times 448$ pixels. Its image and text projections each have 128 dimensions. Text limits are 512 tokens for BioViL-T, 256 for BiomedCLIP, 77 for CLIP and MONET, and 64 for SigLIP-2 and MedSigLIP-448. The panel checkpoints remain frozen. The separate head experiments are described below.

\paragraph{Differences Across the Model Panel.} At the primary conjunction, the 3 highest Recall@1 folds are MedSigLIP-448, BioViL-T, and BiomedCLIP on each dataset. The largest general-domain fold is 1.76. The maximum differences between the 2 CLIP sizes and between SigLIP-2 and CLIP ViT-L/14 are 0.49 and 0.52. MONET's folds range from 1.05 to 1.10. These observations describe the evaluated checkpoints and do not identify which pretraining choice causes their differences.

\paragraph{Parameter Counts.} The counts in \tableref{tab:panelfull} include both encoders. Shared parameters are counted once, and nonparameter buffers are excluded. BioViL-T has 27.3 million image-encoder parameters and 110.2 million text-encoder parameters, including its text projection and masked-language-model head.

\begin{table*}[t]
\floatconts
  {tab:panelfull}
  {\caption{Eight vision-language models and 2 baselines. Each vision-language parameter count includes both encoders.}}
  {\scriptsize
   \setlength{\tabcolsep}{4pt}
   \begin{tabular}{@{}l>{\ttfamily\raggedright\arraybackslash}p{1.95in}rrrl@{}}
   \toprule
   Release & Hugging Face identifier & Params (M) & Input & Embed & Role in the panel \\
   \midrule
   CLIP ViT-B/32 & openai/\allowbreak clip-\allowbreak vit-\allowbreak base-\allowbreak patch32 & 151.3 & $224 \times 224$ & 512 & general domain, small \\
   CLIP ViT-L/14 & openai/\allowbreak clip-\allowbreak vit-\allowbreak large-\allowbreak patch14 & 427.6 & $224 \times 224$ & 768 & general domain, large \\
   SigLIP-2 L/16 & google/\allowbreak siglip2-\allowbreak large-\allowbreak patch16-\allowbreak 512 & 882.3 & $512 \times 512$ & 1024 & training objective \\
   PubMedCLIP & flaviagiammarino/\allowbreak pubmed-\allowbreak clip-\allowbreak vit-\allowbreak base-\allowbreak patch32 & 151.3 & $224 \times 224$ & 512 & biomedical, small \\
   BiomedCLIP & microsoft/\allowbreak BiomedCLIP-\allowbreak PubMedBERT\_\allowbreak 256-\allowbreak vit\_\allowbreak base\_\allowbreak patch16\_\allowbreak 224 & 195.9 & $224 \times 224$ & 512 & biomedical \\
   MedSigLIP-448 & google/\allowbreak medsiglip-\allowbreak 448 & 878.3 & $448 \times 448$ & 1152 & medical, high resolution \\
   BioViL-T & microsoft/\allowbreak BiomedVLP-\allowbreak BioViL-\allowbreak T & 137.6 & $448 \times 448$ & 128 & chest radiology \\
   MONET & chanwkim/\allowbreak monet & 427.6 & $224 \times 224$ & 768 & off-domain medical control \\
   \midrule
   Published metadata & N/A & N/A & N/A & N/A & record linkage, no model \\
   14-label classifier & torchxrayvision densenet121-\allowbreak res224-\allowbreak all & 7.0 & $224 \times 224$ & 12 & label vector, no shared space \\
   \bottomrule
   \end{tabular}}
\end{table*}

\subsection{Attribute and Label Baselines}
\label{apd:nomodel}

\paragraph{Record-Linkage and Label Baselines.} The record-linkage adversary matches the attributes of the primary conjunction, with no image content of any kind. Matching reports receive a score of 1, other reports receive 0, and a deterministic uniform draw smaller than $10^{-3}$ breaks ties. The margin between the 2 largest scores tends to decrease with matching-set size. Margin selection favors smaller sets, without producing the same ordering as explicit smallest-pool selection. Its observed precision at a budget of 100 is 11 to 45 times its Recall@1. The label baseline is the TorchXRayVision \texttt{densenet121-res224-all} checkpoint at $224 \times 224$ pixels. Its outputs cover 12 of the 14 findings, since the checkpoint has no counterpart for no finding or for support devices. Where several outputs name 1 finding, we take the strongest. We compute cosine similarity between these 12 probabilities and each report's corresponding labels, without thresholding. The baselines use the same query-selection rule, budgets, and patient bootstrap as the vision-language models.

\begin{table}[t]
\floatconts
  {tab:nomodel}
  {\caption{Attribute and label baselines at $k=1{,}000$, percent. Boldface marks the maximum.}}
  {\scriptsize
   \setlength{\tabcolsep}{4pt}
   \begin{tabular}{@{}llrrr@{}}
   \toprule
   & & \multicolumn{2}{c}{Unconditioned} & Conditioned \\
   \cmidrule(lr){3-4}
   Adversary & Dataset & R@1$\downarrow$ & Precision$\downarrow$ & Precision$\downarrow$ \\
   \midrule
   Metadata & MIMIC & 0.18 & 1.11 & N/A \\
    & CheXpert & 0.09 & 0.30 & N/A \\
    & ReXGrad. & \textbf{0.49} & \textbf{4.37} & N/A \\
   \midrule
   Classifier & MIMIC & 0.00 & 0.00 & 0.41 \\
    & CheXpert & 0.01 & 0.10 & 0.38 \\
    & ReXGrad. & 0.00 & 0.00 & \textbf{1.91} \\
   \bottomrule
   \end{tabular}}
\end{table}

The record-linkage adversary has no conditioned column in \tableref{tab:nomodel}, because conditioning on the same matched attributes would remove all attribute-based variation among the remaining candidates.

\paragraph{Matched Comparison with Acquisition Year.} The year-conditioned comparison uses 25{,}542 shared eligible queries on ReXGradient-160K. At $k=100$, MedSigLIP-448 with all 4 attributes reaches 99.05\% precision, compared with 55.90\% for the smallest-pool attribute policy and 1.97\% for the model alone. The paired differences are 43.14 and 97.08 percentage points, respectively, with adjusted $p=0.0013$ for both. On the queries selected by the combined attack within each resample, the attribute policy reaches 17.97\% and the unconditioned model reaches 6.52\%. This comparison uses the same 4 attributes for the metadata and combined policies. The uniform-guess expectation on the observed selected queries is 19.91\%.

\section{Linkage Estimation and Report Transformations}
\label{apd:measurement}

\begin{figure}[t]
\floatconts
  {fig:reversal}
  {\caption{(a) Dotted is unconditioned chance. (b) Blue is clinical.}}
  {\includegraphics[width=3.18in]{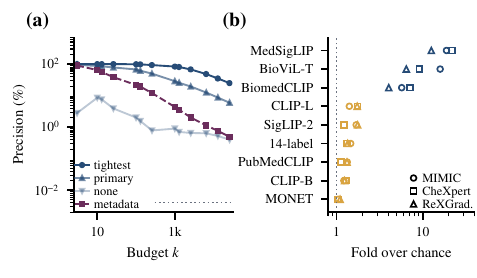}}
\end{figure}

\begin{figure}[t]
\floatconts
  {fig:frontiers}
  {\caption{Primary conjunction, $k=1{,}000$. Dotted is chance AUROC.}}
  {\includegraphics[width=3.18in]{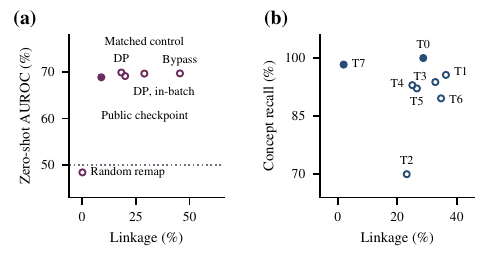}}
\end{figure}

\figureref{fig:reversal}a uses MedSigLIP-448 on ReXGradient-160K. Panel (b) compares Recall@1 folds under the primary conjunction across all 3 datasets. In \figureref{fig:frontiers}, panel (a) compares BioViL-T variants on MIMIC-CXR, with a dotted line at chance AUROC. Panel (b) compares MedSigLIP-448 report transformations on ReXGradient-160K.

\subsection{The Ranking Procedure}
\label{apd:instrument}

\begin{algorithm2e}[t]
\caption{Selective Linkage Precision at an Operating Point}
\label{alg:instrument}
\SetAlgoLined
\DontPrintSemicolon
\KwIn{released encoders $f_{\mathrm{img}}, f_{\mathrm{txt}}$; a dataset's held-out image-report pairs; published attributes $A$; budget $k$}
\KwOut{bootstrap mean precision with uncertainty, observed fold $\phi(k)$, and singleton count $u$}
$\mathcal{Q} \gets \varnothing$; \quad $u \gets 0$ \tcp*{ranked queries; queries identified by attributes alone}
\ForEach{query image $i$}{
  $\mathcal{C}_i \gets \{\, r : a(r) = a(i) \,\}$ \tcp*{from the published attributes}
  $N_i \gets |\mathcal{C}_i|$\;
  \lIf{$N_i = 1$}{$u \gets u + 1$; \textbf{continue}}
  $s(r) \gets \cos\!\left(f_{\mathrm{img}}(x_i), f_{\mathrm{txt}}(r)\right)$ for $r \in \mathcal{C}_i$\;
  $s_{(1)}, s_{(2)} \gets$ the 2 largest values of $s$; \quad $\hat r_i \gets \arg\max_r s(r)$\;
  $m_i \gets s_{(1)} - s_{(2)}$ \tcp*{computable with no access to the truth}
  $c_i \gets \mathbf{1}\!\left[s(r_i)=s_{(1)}\right]$ \tcp*{the true report may tie at the maximum}
  $\mathcal{Q} \gets \mathcal{Q} \cup \{(m_i, c_i, N_i, \mathrm{patient}(i))\}$\;
}
\SetKwFunction{Est}{estimate}
\SetKwProg{Fn}{Function}{:}{}
\Fn{\Est{$\mathcal{Q}$}}{
  $k' \gets \min(k,|\mathcal{Q}|)$\;
  $\mathcal{S} \gets$ the $k'$ entries of $\mathcal{Q}$ with the largest $m_i$\;
  \KwRet $\left(\tfrac{1}{k'}\sum_{i \in \mathcal{S}} c_i,\ \ \tfrac{1}{k'}\sum_{i \in \mathcal{S}} \tfrac{1}{N_i}\right)$ \tcp*{chance is taken over the selected queries}
}
$(P, \bar{p}) \gets$ \Est{$\mathcal{Q}$}; \quad $\phi \gets P / \bar{p}$\;
\For{$b \gets 1$ \KwTo $1{,}000$}{
  $\mathcal{Q}^{b} \gets$ resample the patients of $\mathcal{Q}$ with replacement\;
  $(P^{b}, \bar{p}^{\,b}) \gets$ \Est{$\mathcal{Q}^{b}$} \tcp*{re-select inside every resample}
}
\KwRet mean, standard deviation, and 95\% percentile interval of $\{P^{b}\}$; observed fold $\phi$; singleton count $u$\;
\end{algorithm2e}

\paragraph{Extraction and Scoring.} Embeddings use single precision and batches of 64. Scoring computes cosine similarities between normalized image embeddings and distinct candidate-report embeddings. Sorting queries once by margin produces the budget curve. \algorithmref{alg:instrument} specifies the primary complete-collection evaluation and its patient bootstrap.

\paragraph{Paired Comparisons.} For a difference in selective precision, we resample patients jointly across the 2 policies and repeat each policy's query selection within every resample. We use 1{,}000 replicates with seed 0. Let $\hat\Delta$ be the observed difference and $\Delta_b$ its value in replicate $b$. The two-sided bootstrap $p$ is the fraction of finite replicates for which $|\Delta_b-\hat\Delta|\geq|\hat\Delta|$. Its minimum is the reciprocal of the number of finite replicates. We then apply the stated false discovery rate correction. These comparisons account for shared queries and patient clustering. They do not remove differences between the training procedures.

\paragraph{The Pathology-Matched Pool.} This control selects 10{,}000 candidate reports with the fewest label disagreements with the query, counting disagreements only over findings labeled in both. We retain the true report. The candidates are not required to have identical labels. Because this construction uses the query's ground-truth findings, it evaluates retrieval among similar reports without representing the assumed attacker.

\subsection{Missing Reports, Attribute Errors, and Collection Size}
\label{apd:costsdetail}

\tableref{tab:costs} evaluates missing correct reports, incorrect attributes, and larger collections. For missing reports, we score each query with and without its true report, then assign it to 1 condition using a deterministic draw from its identifier with seed 0. Assignments are nested across absent-report fractions. Queries without a second candidate after removal are excluded because their margin is undefined. We recompute selective precision and the fraction of selected queries with absent reports. We also evaluate whether the margin distinguishes the present and absent conditions using the area under the receiver operating characteristic curve. For incorrect attributes, each query attribute is independently perturbed at the stated rate using an identifier-seeded draw. Age band changes by 1 band, year by 1 year, and categorical values change to another category in the collection. Filtering then uses these perturbed values, which can exclude the true report. For larger collections, we use either the union of the 3 test splits or every study in the same dataset, including training and validation studies. The query images remain from the test split, and paired comparisons use shared eligible queries.

\begin{table*}[t]
\floatconts
  {tab:costs}
  {\caption{Precision at $k=100$ (percent), MedSigLIP-448 at the primary conjunction. Pool is the median candidate pool on ReXGradient-160K.}}
  {\scriptsize
   \setlength{\tabcolsep}{3pt}
   \begin{tabular}{@{}llrrr@{}}
   \toprule
   Condition & Pool & MIMIC$\downarrow$ & CheXpert$\downarrow$ & ReXGrad.$\downarrow$ \\
   \midrule
   As evaluated above & 327 & 35.38 & 9.48 & 67.68 \\
   \midrule
   \multicolumn{5}{@{}l}{\textbf{Correct report absent from}}\\
   \quad a quarter of the queries & N/A & 27.82 & 5.23 & 55.97 \\
   \quad half of the queries & N/A & 17.66 & 3.11 & 38.52 \\
   \midrule
   \multicolumn{5}{@{}l}{\textbf{Attribute error rate of}}\\
   \quad 5\% & 318 & 29.10 & 7.36 & 41.26 \\
   \quad 10\% & 316 & 21.94 & 6.69 & 29.65 \\
   \quad 20\% & 298 & 12.53 & 6.48 & 11.79 \\
   \midrule
   \multicolumn{5}{@{}l}{\textbf{Candidate collection of}}\\
   \quad 3 test splits & 866 & 28.45 & 9.48 & 53.12 \\
   \quad the whole dataset & 1711 & 14.38 & N/A & 37.23 \\
   \bottomrule
   \end{tabular}}
\end{table*}

Keyword patterns detect mentions of sex, age, and projection in 45\%, 0\%, and 87\% of MIMIC-CXR reports; 60\%, 24\%, and 58\% of CheXpert Plus reports; and 4\%, 5\%, and 13\% of ReXGradient-160K reports, respectively. These are text-pattern detection rates, not verified attribute values or lower bounds on metadata availability.

\figureref{fig:assumptions} plots the 3 perturbations on ReXGradient-160K. Each changes 1 assumption while retaining the primary conjunction and query budget.

\begin{figure*}[t]
\floatconts
  {fig:assumptions}
  {\caption{MedSigLIP-448 on ReXGradient-160K, primary conjunction, $k=100$: (a) missing reports, (b) attribute errors, and (c) larger collections. Points are bootstrap means.}}
  {\includegraphics[width=6.5in]{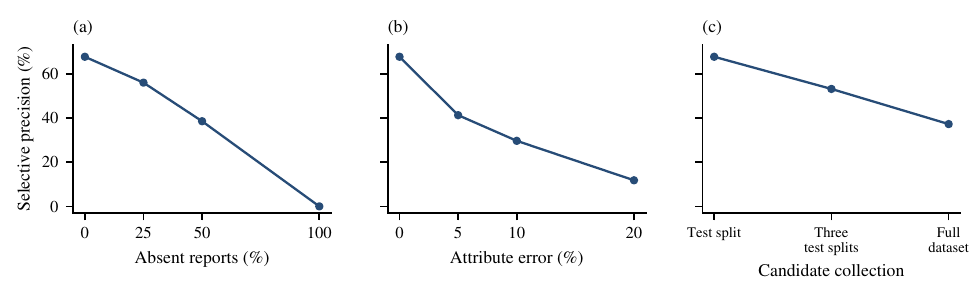}}
\end{figure*}

\subsection{Report Covariates}
\label{apd:map}

\paragraph{Report Covariates and Their Computation.} Report length counts whitespace-separated tokens. Section count is the number of nonempty sections recognized by the parser, and positive findings counts asserted labels. Binary patterns detect laterality (left, right, or bilateral), devices (tube, catheter, line, pacemaker, stent, drain, wire, or valve), comparisons (compar, prior, previous, interval change, or since), and measurements (a number followed by cm or mm). These are case-insensitive whole-word patterns. They differ from the transformation strip groups in \appendixref{apd:transforms}. Rare finding indicates an asserted finding whose prevalence among labeled evaluation studies is below the fixed 2\% threshold. The placeholder count detects runs of at least 3 underscores. ReXGradient-160K has only 1 positive query for this covariate. We treat its slope descriptively and do not interpret its significance.

\paragraph{Criteria for Inclusion in the Agreement Count.} We include models with at least 10 correct query matches and a Recall@1 fold of at least 2.0. These criteria select BiomedCLIP, BioViL-T, and MedSigLIP-448 on all 6 combinations of dataset and operating point. Models near chance can recover many queries when pools are small. Therefore, we require both criteria. We report sign agreement for the 8 covariates other than the sparse placeholder count.

\begin{table*}[t]
\floatconts
  {tab:map}
  {\caption{Standardized slope of each report covariate on the margin, $10^{-3}$ units, primary conjunction. Boldface marks significance after false discovery rate correction.}}
  {\scriptsize
   \setlength{\tabcolsep}{3pt}
   \begin{tabular}{@{}lrrrrrrrrr@{}}
   \toprule
   & \multicolumn{3}{c}{MIMIC-CXR} & \multicolumn{3}{c}{CheXpert Plus} & \multicolumn{3}{c}{ReXGradient-160K} \\
   \cmidrule(lr){2-4}\cmidrule(lr){5-7}\cmidrule(lr){8-10}
   Report covariate & BioViL-T & \shortstack[r]{MedSigLIP-\\448} & \shortstack[r]{Biomed\\CLIP} & BioViL-T & \shortstack[r]{MedSigLIP-\\448} & \shortstack[r]{Biomed\\CLIP} & BioViL-T & \shortstack[r]{MedSigLIP-\\448} & \shortstack[r]{Biomed\\CLIP} \\
   \midrule
Report length in tokens & \textbf{$+$0.59} & \textbf{$+$0.97} & $+$0.06 & $+$0.10 & $-$0.02 & $-$0.05 & \textbf{$+$2.41} & \textbf{$+$2.45} & \textbf{$+$0.53} \\
Section count & \textbf{$+$0.44} & \textbf{$-$0.37} & \textbf{$+$0.15} & $-$0.31 & $+$0.05 & \textbf{$+$0.11} & \textbf{$+$1.36} & \textbf{$+$0.74} & $+$0.07 \\
Positive findings & \textbf{$+$0.35} & \textbf{$+$1.33} & \textbf{$-$0.30} & $-$0.39 & \textbf{$+$0.16} & \textbf{$-$0.30} & \textbf{$+$2.72} & \textbf{$+$3.03} & \textbf{$+$0.43} \\
States laterality & \textbf{$+$0.40} & \textbf{$+$1.25} & $+$0.03 & $-$0.12 & $+$0.13 & $-$0.06 & \textbf{$+$2.25} & \textbf{$+$2.40} & \textbf{$+$0.50} \\
Mentions a device & \textbf{$+$0.44} & \textbf{$+$1.01} & \textbf{$-$0.28} & \textbf{$-$0.55} & \textbf{$-$0.26} & \textbf{$-$0.22} & \textbf{$+$1.84} & \textbf{$+$1.52} & \textbf{$+$0.30} \\
Compares with a prior & $-$0.10 & \textbf{$+$0.67} & $+$0.04 & $+$0.15 & $+$0.08 & $+$0.03 & \textbf{$+$1.52} & \textbf{$+$1.42} & \textbf{$+$0.34} \\
Contains a measurement & \textbf{$+$0.51} & \textbf{$+$0.23} & \textbf{$-$0.26} & $-$0.10 & $-$0.13 & \textbf{$-$0.10} & $+$0.38 & \textbf{$+$0.39} & $+$0.08 \\
Contains a rare finding & $+$0.18 & $+$0.08 & $+$0.05 & $+$0.31 & $+$0.09 & $+$0.04 & \textbf{$+$0.91} & \textbf{$+$0.61} & $+$0.08 \\
De-identification placeholders & \textbf{$+$0.40} & \textbf{$+$0.48} & $+$0.02 & $+$0.03 & \textbf{$-$0.14} & $-$0.01 & $+$0.02 & $+$0.14 & $+$0.02 \\
   \bottomrule
   \end{tabular}}
\end{table*}

\begin{figure}[t]
\floatconts
  {fig:map}
  {\caption{Covariate slopes. An asterisk marks significance.}}
  {\includegraphics[width=3.18in]{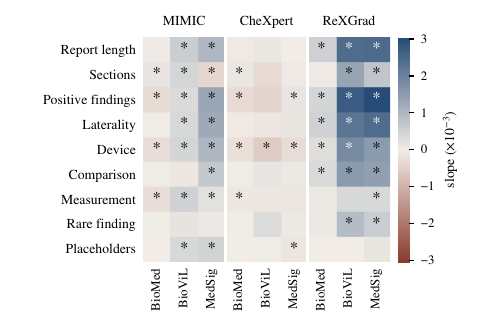}}
\end{figure}

\figureref{fig:map} plots the report-covariate slopes. \tableref{tab:map} reports the standardized slope of every covariate on the margin, on all 3 datasets. Each fit is an ordinary least-squares regression of the margin on the standardized covariate, with a cluster-robust standard error on the patient. The patient-clustered standard error allows arbitrary correlation among observations from the same patient. We use the same procedure for all 837 reported slopes.

For the 8 covariates used in the agreement summary, 3 agree in margin-slope sign on MIMIC-CXR at both operating points. CheXpert Plus has agreement for 3 without conditioning and 4 with conditioning. ReXGradient-160K has agreement for all 8 at both operating points.

\subsection{The Joint Covariate Fit}
\label{apd:adjusted}

The 9 covariates can be correlated with each other and with the candidate-pool size. We fit them jointly in 1 ordinary least-squares model, together with the logarithm of the candidate-pool size as a tenth term, with standard errors clustered by patient. We fit 2 outcomes, the margin and a binary indicator of a correct link, because a larger margin does not necessarily imply a correct match. The variance inflation factor measures how strongly each predictor is associated with the remaining predictors.

\tableref{tab:adjusted} reports both outcomes for MedSigLIP-448 on ReXGradient-160K at the primary conjunction. The fit includes the placeholder count, but we omit its coefficient from the table because only 1 query contains a detected placeholder. The remaining coefficients describe adjusted associations. We do not evaluate their predictive accuracy.

\begin{table*}[t]
\floatconts
  {tab:adjusted}
  {\caption{Joint covariate slopes per standard deviation and adjusted $p$. VIF is the variance inflation factor. MedSigLIP-448, ReXGradient-160K, primary conjunction.}}
  {\scriptsize
   \setlength{\tabcolsep}{3pt}
   \begin{tabular}{@{}lrrrrr@{}}
   \toprule
   Covariate & Correct link (points) & $p$ & Margin ($10^{-3}$) & $p$ & VIF \\
   \midrule
   States laterality & $+$1.84 & $<$0.001 & $+$0.94 & $<$0.001 & 1.47 \\
   Mentions a device & $+$1.61 & $<$0.001 & $+$0.66 & $<$0.001 & 1.45 \\
   Positive findings & $+$2.60 & $<$0.001 & $+$2.13 & $<$0.001 & 1.49 \\
   Section count & $+$0.86 & $<$0.001 & $+$0.30 & 0.009 & 1.05 \\
   Compares with a prior & $+$0.58 & 0.002 & $+$0.53 & $<$0.001 & 1.12 \\
   Contains a rare finding & $+$0.70 & 0.009 & $+$0.04 & 0.786 & 1.03 \\
   Contains a measurement & $-$0.09 & 0.696 & $-$0.56 & $<$0.001 & 1.28 \\
   Report length in tokens & $-$0.23 & 0.301 & $+$1.06 & $<$0.001 & 1.66 \\
   \textbf{Log candidate-pool size} & $-$3.91 & $<$0.001 & $-$2.66 & $<$0.001 & 1.04 \\
   \bottomrule
   \end{tabular}}
\end{table*}

\figureref{fig:adjusted} compares the correct-link coefficients on their common standardized scale. The pool-size coefficient has the largest magnitude and is negative. Positive findings have the largest positive coefficient.

\begin{figure}[t]
\floatconts
  {fig:adjusted}
  {\caption{Adjusted correct-link coefficients, MedSigLIP-448, ReXGradient-160K.}}
  {\includegraphics[width=3.18in]{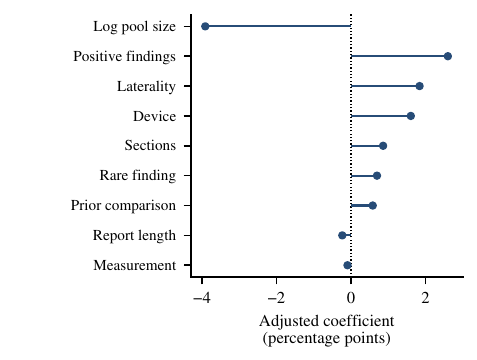}}
\end{figure}

\subsection{Report Transformations}
\label{apd:transforms}

\paragraph{Strip Groups.} The measurement group matches dimensions in centimeters or millimeters in 1, 2, or 3 axes, catheter and tube gauges in French or gauge, bare percentages, and a cardiothoracic ratio with its value. The date group matches numeric dates in 3 orders, dates written with the month name, clock times, and ages written as a number of years old. The device group contains 52 terms for airway and enteric tubes, central and peripheral lines, chest drains, cardiac devices and leads, surgical hardware, mechanical support, and stents and valves. Each term is matched as a whole word with an optional plural. Each group is implemented as a single term list.

\paragraph{The Label Template.} T7 replaces the narrative with text derived from the study's 14 labels. Under a \texttt{FINDINGS:} header, it names asserted findings or states no acute cardiopulmonary abnormality. An optional line names explicitly negated findings. Under an \texttt{IMPRESSION:} header, it repeats the asserted set or states no acute abnormality. Residual linkage under T7 combines this coarse label representation with the published attributes and the model scores.

\paragraph{Rewriting with a Language Model.} T6 rewrites each report with Qwen3.5-122B-A10B, using the local GPTQ-Int4 build at temperature 0, top-$p$ 1.0, seed 0, a 512-token limit, and the model's chat template with reasoning disabled. All report inference runs locally. The same model and settings produce the 14 labels for ReXGradient-160K and relabel the original and transformed reports for label agreement and concept recall. The 6 additional content axes use text patterns and no language model. The rewrite prompt is

\begin{quote}\small
``You are a radiology editor. Rewrite the chest radiograph report you are given into one fixed, canonical style. Keep every clinical finding, its laterality and its severity. Remove wording that identifies the individual report: measurements, dates, times, comparisons to prior studies, technique and positioning notes, hedging phrases, and any turn of phrase specific to the dictating radiologist. Write exactly two sections, FINDINGS: and IMPRESSION:, in plain declarative sentences. Output the report only.''
\end{quote}

\noindent followed by the report text. The labeling prompt is

\begin{quote}\small
``You label chest radiograph reports. For each of the 14 findings below, answer 1 if the report states the finding is present, 0 if it states the finding is absent or is uncertain about it, and 3 if the report does not mention it at all. Answer with a JSON object only, with one key per finding and no other text.''
\end{quote}

\noindent followed by the 14 finding names, the report text, and a \texttt{JSON:} cue. An unmentioned finding is excluded from this labeler's comparisons. Its treatment of uncertainty differs from the supplied MIMIC-CXR and CheXpert Plus labels described in \appendixref{apd:datasets}.

\paragraph{Computing the Content Measures.} \tableref{tab:transforms} compares linkage and label agreement under the definitions in \sectionref{sec:defense}. The fixed sample contains 2{,}000 studies per dataset, drawn once with seed 11. Label agreement is first averaged over comparable findings within each study. Concept recall is computed only for studies with at least 1 asserted finding in the original report. We compute both measures per study and bootstrap over patients because the target is the per-study mean.

\begin{table*}[t]
\floatconts
  {tab:transforms}
  {\caption{Report transformations at the primary conjunction, percent. Precision is MedSigLIP-448 at $k=1{,}000$. Green marks intervals that fall below the shaded T0 row.}}
  {\scriptsize
   \setlength{\tabcolsep}{2pt}
   \begin{tabular}{@{}llrrrrrrrrr@{}}
   \toprule
   & & \multicolumn{3}{c}{MIMIC-CXR} & \multicolumn{3}{c}{CheXpert Plus} & \multicolumn{3}{c}{ReXGradient-160K} \\
   \cmidrule(lr){3-5}\cmidrule(lr){6-8}\cmidrule(lr){9-11}
    & Transformation & Precision & \shortstack[r]{Label\\agreement} & \shortstack[r]{Concept\\recall} & Precision & \shortstack[r]{Label\\agreement} & \shortstack[r]{Concept\\recall} & Precision & \shortstack[r]{Label\\agreement} & \shortstack[r]{Concept\\recall} \\
   \midrule
   \rowcolor{gray!12} T0 & Untransformed report & 14.94 & 100.00 & 100.00 & 6.42 & 100.00 & 100.00 & 28.74 & 100.00 & 100.00 \\
   T1 & Findings and impression only & 22.78 & 98.26 & 94.03 & 9.17 & 97.23 & 92.23 & 36.37 & 98.69 & 95.65 \\
   T2 & Impression section only & 18.89 & 94.97 & 74.39 & 8.97 & 96.17 & 84.69 & 23.20 & 93.89 & 69.98 \\
   T3 & Measurements and dates stripped & 15.87 & 99.39 & 97.20 & 7.89 & 98.31 & 97.55 & 32.80 & 99.33 & 93.80 \\
   T4 & Device mentions stripped & 13.86 & 98.98 & 93.57 & 4.58 & 98.20 & 94.64 & 25.07 & 99.15 & 92.99 \\
   T5 & Measurements, dates, and devices & 14.35 & 99.03 & 93.76 & 5.29 & 97.84 & 94.57 & 26.61 & 99.15 & 92.15 \\
   T6 & Language-model rewrite & 26.13 & 97.48 & 93.90 & 9.44 & 96.29 & 92.95 & 34.72 & 97.74 & 89.58 \\
   T7 & Template from the 14 labels & \cellcolor{green!22}0.30 & 86.24 & 64.22 & \cellcolor{green!22}0.10 & 88.41 & 76.20 & \cellcolor{green!22}1.99 & 98.95 & 98.32 \\
   \bottomrule
   \end{tabular}}
\end{table*}

The same language-model labeling policy is used for these agreement measures on all datasets. The supplied labels used to construct T7 differ from that policy on MIMIC-CXR and CheXpert Plus. Therefore, T7 agreement also reflects differences between the template's input labels and the evaluation labeler.

\paragraph{Six Additional Content Measures.} Agreement over 14 findings does not assess the details surrounding those findings, and the labeler merges uncertainty with absence. We measure 6 additional types of mention with text patterns. Laterality, measurements, devices, and prior comparisons use the patterns defined in \appendixref{apd:map}. Severity matches qualifiers such as mild, moderate, or severe. Hedging matches qualifiers such as possible, probable, or cannot exclude. For each axis, the denominator contains sampled reports with an original mention. The numerator contains those whose transformed version still has a mention on that axis. These measures require no language-model labels. They do not verify that the mentioned detail remains correct.

\tableref{tab:axes} reports each axis by transformation and dataset. Measurement mentions disappear after measurement stripping. Device detection can remain positive after device stripping because it also matches generic words such as tube and line, while the strip includes specific device phrases. These pattern counts do not assess clinical adequacy.

\begin{table*}[t]
\floatconts
  {tab:axes}
  {\caption{Content retained on 6 axes and finding-label agreement, in percent. Each axis is the fraction of reports with an original mention that retain a mention after transformation.}}
  {\scriptsize
   \setlength{\tabcolsep}{2.6pt}
   \begin{tabular}{@{}llrrrrrrr@{}}
   \toprule
   & & \multicolumn{6}{c}{Reports retaining a mention on each axis} & \\
   \cmidrule(lr){3-8}
   Transformation & Dataset & Laterality & Severity & \shortstack[r]{Measure-\\ment} & Device & \shortstack[r]{Prior\\study} & Hedging & Agreement \\
   \midrule
   Untransformed report & MIMIC-CXR & 100.00 & 100.00 & 100.00 & 100.00 & 100.00 & 100.00 & 100.00 \\
    & CheXpert Plus & 100.00 & 100.00 & 100.00 & 100.00 & 100.00 & 100.00 & 100.00 \\
    & ReXGradient-160K & 100.00 & 100.00 & 100.00 & 100.00 & 100.00 & 100.00 & 100.00 \\
   \midrule
   Findings and impression only & MIMIC-CXR & 94.70 & 95.92 & 95.09 & 94.64 & 75.40 & 89.83 & 98.26 \\
    & CheXpert Plus & 99.44 & 79.04 & 97.92 & 97.20 & 87.19 & 57.35 & 97.23 \\
    & ReXGradient-160K & 92.23 & 99.17 & 100.00 & 97.76 & 67.85 & 95.98 & 98.69 \\
   \midrule
   Impression section only & MIMIC-CXR & 71.43 & 73.17 & 59.16 & 68.12 & 50.46 & 63.15 & 94.97 \\
    & CheXpert Plus & 95.40 & 72.51 & 71.08 & 86.31 & 71.85 & 47.66 & 96.17 \\
    & ReXGradient-160K & 62.50 & 56.13 & 51.18 & 52.73 & 28.80 & 67.84 & 93.89 \\
   \midrule
   Measurements and dates stripped & MIMIC-CXR & 100.00 & 100.00 & 0.00 & 100.00 & 100.00 & 100.00 & 99.39 \\
    & CheXpert Plus & 100.00 & 100.00 & 0.00 & 100.00 & 100.00 & 98.87 & 98.31 \\
    & ReXGradient-160K & 100.00 & 100.00 & 0.00 & 100.00 & 100.00 & 98.31 & 99.33 \\
   \midrule
   Device mentions stripped & MIMIC-CXR & 100.00 & 100.00 & 100.00 & 53.09 & 100.00 & 100.00 & 98.98 \\
    & CheXpert Plus & 100.00 & 100.00 & 100.00 & 46.70 & 100.00 & 100.00 & 98.20 \\
    & ReXGradient-160K & 100.00 & 100.00 & 100.00 & 40.42 & 100.00 & 100.00 & 99.15 \\
   \midrule
   Measurements, dates, and devices & MIMIC-CXR & 100.00 & 100.00 & 0.00 & 53.09 & 100.00 & 100.00 & 99.03 \\
    & CheXpert Plus & 100.00 & 100.00 & 0.00 & 46.70 & 100.00 & 98.87 & 97.84 \\
    & ReXGradient-160K & 100.00 & 100.00 & 0.00 & 40.42 & 100.00 & 98.31 & 99.15 \\
   \midrule
   Language-model rewrite & MIMIC-CXR & 96.62 & 94.29 & 88.37 & 94.90 & 7.97 & 48.91 & 97.48 \\
    & CheXpert Plus & 99.43 & 77.43 & 93.68 & 94.61 & 15.00 & 41.06 & 96.29 \\
    & ReXGradient-160K & 91.64 & 93.14 & 89.33 & 96.98 & 11.19 & 59.43 & 97.74 \\
   \midrule
   Template from the 14 labels & MIMIC-CXR & 0.00 & 0.00 & 0.00 & 0.00 & 0.00 & 0.00 & 86.24 \\
    & CheXpert Plus & 0.00 & 0.00 & 0.00 & 0.00 & 0.00 & 0.00 & 88.41 \\
    & ReXGradient-160K & 0.00 & 0.00 & 0.00 & 0.00 & 0.00 & 0.00 & 98.95 \\
   \bottomrule
   \end{tabular}}
\end{table*}

\section{Head Training and Privacy Accounting}
\label{apd:training}

\subsection{Differentially Private Finetuning of the Projection Heads}
\label{apd:dp}

\paragraph{Parameters Under the Guarantee.} We train only the 2 projection heads, which have 263{,}424 parameters in total, and both encoders of BioViL-T remain frozen. Therefore, the privacy accounting applies to the parameters of the projection heads that are updated during finetuning and to no others. The private dataset is MIMIC-CXR's training split, and 1 record is 1 image.

\paragraph{The Mechanism.} Training uses differentially private (DP) stochastic gradient descent \citep{DPSGD}. We clip the gradient of each example to a maximum $L_2$ norm of 1.5, add Gaussian noise to the summed clipped gradients, and update the heads with AdamW from the noisy average. We use Opacus \citep{Yousefpour2021Opacus} to calibrate the noise multiplier for a target $\varepsilon$ of 2.0 at $\delta = 6 \times 10^{-6}$. The planned run has 150 epochs with logical batches of 128 sampled from the 154{,}611 training images. We assemble the logical batch from physical batches of 64, because per-example gradients for all 128 examples do not fit in memory at once. Gradient accumulation leaves the computed gradient and the privacy accounting unchanged. The heads contain no normalization layer that couples examples within a batch, since a per-sample gradient through such a layer would not be a function of that sample alone.

\paragraph{The Decomposable Objective.} The standard per-example clipping argument assumes a loss of the form $L(\theta; D) = \sum_i \ell(\theta; x_i)$. Each term depends on the parameters and a single private example. Symmetric in-batch InfoNCE does not satisfy this assumption. Example $i$ appears in a softmax whose denominator contains every other example in the batch. Therefore, removing 1 example changes every other example's term. We replace the in-batch negatives with a fixed bank of 4{,}096 validation cases drawn under the global seed. The validation patients are disjoint from the training patients and outside the private dataset. Each example's loss contains its image-to-text and text-to-image terms computed using its true pair and that bank. The loss is returned without reduction. The loss then depends on 1 private example, the parameters, and fixed non-private inputs, satisfying the required decomposition.

\paragraph{The Achieved Guarantee.} The final accountant value after 150 epochs is $\varepsilon = 1.991574$, reported as 1.99, at $\delta = 6 \times 10^{-6}$ per training image. We use this full-run bound for the selected checkpoint. The private run selects the lowest validation selective precision, attained after 18 epochs. The matched non-private control and the bypass select the highest validation selective precision. These opposite selection criteria confound attribution of the test-set difference to clipping and noise. The in-batch comparator has the same nominal accountant value. Its nondecomposable objective invalidates that accounting argument.

\paragraph{Limits of the Accounting.} For a patient contributing $q$ images, group privacy gives the conservative bound $(q\varepsilon,\; q e^{(q-1)\varepsilon}\delta)$ from the example-level guarantee. The accounting uses the whole-pool maximum of $q=124$, compared with 122 in the training split. The resulting bound is $\varepsilon'=246.96$ with $\delta'$ of order $10^{103}$. A $\delta'$ above 1 provides no constraint, making this patient-level bound vacuous. Using the maximum contribution for every patient further loosens the bound. Training with patient-level sampling and clipping would directly account for patients and require a different training run. The in-batch comparator's reported accountant value does not establish a guarantee because its loss does not satisfy the per-example sensitivity argument. We retain that comparator without a privacy guarantee.

\subsection{Release Variants}
\label{apd:variants}

\begin{table*}[t]
\floatconts
  {tab:dpconfig}
  {\caption{Head variants and the matched control on frozen BioViL-T encoders. The trained variants use AdamW. The random remapping is untrained.}}
  {\scriptsize
   \setlength{\tabcolsep}{3.6pt}
   \begin{tabular}{@{}llllllllll@{}}
   \toprule
   Variant & Negatives & Clip + noise & Learning rate & Weight decay & Batch & Epochs & Clip norm & Target $\varepsilon$ & $\delta$ \\
   \midrule
   DP heads & fixed bank, 4{,}096 & yes & 0.005 & 0.01 & 128 & 150 & 1.5 & 2.0 & $6 \times 10^{-6}$ \\
   DP in-batch & in-batch & yes & 0.005 & 0.01 & 128 & 150 & 1.5 & 2.0 & $6 \times 10^{-6}$ \\
   Matched control & fixed bank, 4{,}096 & no & 0.005 & 0.01 & 128 & 150 & N/A & N/A & N/A \\
   Bypass heads & in-batch & no & 0.005 & 0.01 & 512 & 50 & N/A & N/A & N/A \\
   Random remap & none & no & N/A & N/A & N/A & N/A & N/A & N/A & N/A \\
   \bottomrule
   \end{tabular}}
\end{table*}

\paragraph{Projection Heads and Variant Training.} \tableref{tab:dpconfig} lists every training setting. Each head is a 2-layer perceptron that takes its encoder's output embedding. It has 512 hidden units with a GELU activation \citep{Hendrycks2016Gelu} and a 128-dimensional output, and both outputs are $L_2$ normalized before the cosine. The head takes the public model's output embedding and not an intermediate activation, which is the representation available to an adversary with the public model. The contrastive temperature is 0.07 and the optimizer is AdamW \citep{Loshchilov2019Adamw} throughout.

\paragraph{Variants with No Privacy Mechanism.} The non-private control uses the same head architecture, fixed set of 4{,}096 validation negatives, nominal batch size, learning rate, weight decay, training duration, and seed as the private run. It omits clipping and noise. It also differs in checkpoint selection and uses shuffled fixed-size batches instead of the private run's Poisson sampling. Consequently, this control does not isolate the effect of clipping and noise. The bypass uses batches of 512 for 50 epochs to fit another alignment on public paired data. It represents an attacker with access to the frozen encoder outputs. The random orthogonal variant requires no training. We independently draw an orthogonal map for each encoder by QR decomposition of a Gaussian matrix with the sign convention fixed. Each map preserves norms and inner products within its modality, while the independent maps change the correspondence between modalities.

\paragraph{Utility Measures.} Zero-shot classification uses cosine similarities between the image and 2 fixed prompts for each finding, ``findings consistent with \{finding\}'' and ``no evidence of \{finding\}''. We report the macro area under the receiver operating characteristic curve (AUROC) across the 14 findings. Each finding's AUROC uses a patient bootstrap. The macro estimate resamples the 14 finding-specific AUROCs, describing variation across findings. We fit a separate image-side linear classifier for each finding on frozen image embeddings with a learning rate of 0.001, weight decay of 0.0001, and batches of 512. Training runs for up to 100 epochs, with early stopping after 10 epochs without improvement in that finding's validation AUROC. We select each classifier's checkpoint by that criterion. \tableref{tab:variants} reports all 3 utility measures beside linkage at every budget. Image-side AUROC differs by 0.03 percentage points under the random orthogonal variant. Independently of this estimate, an invertible orthogonal map preserves linear separability. To limit the quadratic cost of text-to-image retrieval, we sample 5{,}000 distinct studies and report the observed number of correct matches. Sampling images could include several images with the same report and make the correct match ambiguous.

\begin{figure}[t]
\floatconts
  {fig:defensepairs}
  {\caption{The 36 paired differences, in points. Gray markers are not significant.}}
  {\includegraphics[width=3.18in]{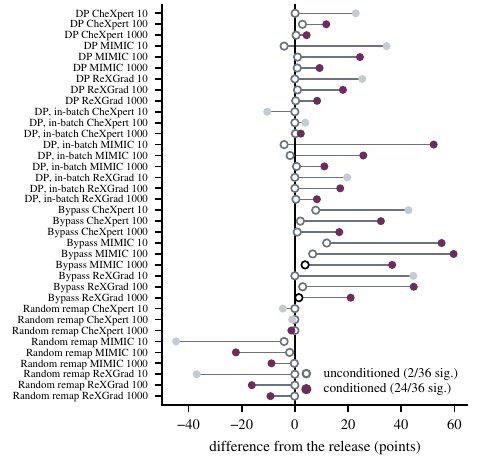}}
\end{figure}

\begin{table*}[t]
\floatconts
  {tab:variants}
  {\caption{Primary conjunction at $k=10$, 100, and 1{,}000, with MIMIC-CXR utility. Text-to-image counts are out of 5{,}000. Other values are percentages. The public checkpoint is shaded.}}
  {\scriptsize
   \setlength{\tabcolsep}{3pt}
   \begin{tabular}{@{}lrrrrrrrrrrrr@{}}
   \toprule
   & \multicolumn{3}{c}{MIMIC-CXR} & \multicolumn{3}{c}{CheXpert Plus} & \multicolumn{3}{c}{ReXGradient-160K} & \multicolumn{3}{c}{Utility} \\
   \cmidrule(lr){2-4}\cmidrule(lr){5-7}\cmidrule(lr){8-10}\cmidrule(lr){11-13}
   Variant & 10 & 100 & 1{,}000 & 10 & 100 & 1{,}000 & 10 & 100 & 1{,}000 & Zero-shot & \shortstack{Text-to-image\\count} & Image \\
   \midrule
   \rowcolor{gray!12} Public checkpoint & 44.68 & 22.18 & 8.96 & 14.16 & 2.01 & 1.40 & 49.09 & 24.10 & 10.37 & 68.86 & 73 & 78.77 \\
   Matched non-private control & 87.67 & 62.89 & 28.97 & 35.54 & 24.45 & 10.76 & 87.55 & 52.16 & 27.00 & 69.66 & 116 & 77.49 \\
   DP heads & 79.17 & 46.66 & 18.26 & 37.06 & 13.82 & 5.82 & 74.45 & 42.16 & 18.74 & 69.89 & 77 & 78.41 \\
   DP heads, in-batch loss & 96.88 & 47.92 & 20.05 & 3.78 & 5.93 & 3.63 & 68.77 & 41.15 & 18.66 & 69.12 & 108 & 79.08 \\
   Bypass heads on public data & 99.86 & 81.87 & 45.55 & 56.86 & 34.35 & 18.11 & 93.63 & 68.80 & 31.35 & 69.70 & 306 & 80.21 \\
   Random orthogonal remapping & 0.00 & 0.00 & 0.16 & 9.63 & 0.96 & 0.10 & 12.17 & 7.90 & 1.20 & 48.38 & 2 & 78.80 \\
   \bottomrule
   \end{tabular}}
\end{table*}

\tableref{tab:variants} reports every variant at every budget. \figureref{fig:defensepairs} shows the 36 paired differences at both operating points.

\figureref{fig:trainingcomparison} compares the public checkpoint, matched control, and private run. The private run's precision is lower than the control's and higher than the public checkpoint's on each dataset. Checkpoint selection and batch sampling differ between the 2 trained runs, as described above.

\begin{figure}[t]
\floatconts
  {fig:trainingcomparison}
  {\caption{BioViL-T training comparison, primary conjunction, $k=1{,}000$.}}
  {\includegraphics[width=3.18in]{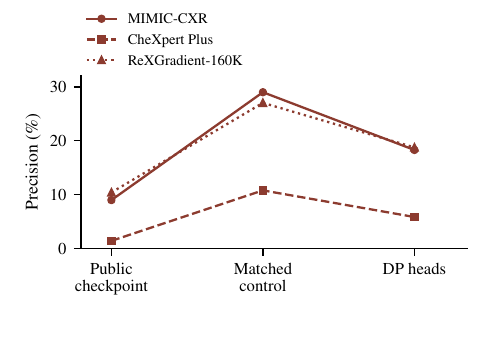}}
\end{figure}

\section{Extended Linkage and Report-Content Results}
\label{apd:extended}

\subsection{Retrieval Metrics}
\label{apd:field}

\paragraph{Metric Definitions.} Recall@$k$ is the share of queries whose true report falls in the top $k$ candidates. The mean reciprocal rank averages $1/\mathrm{rank}$ of the true report over queries. These metrics use each query's ranking of candidate reports. Selective precision additionally ranks the queries by margin and evaluates the highest-ranked subset. Recall@1 equals $P(k)$ at $k = M$, where all eligible queries are included.

\begin{table*}[t]
\floatconts
  {tab:fieldmetrics}
  {\caption{Retrieval bootstrap means $\pm$ standard deviation in percent, all $\downarrow$. MRR is the mean reciprocal rank, and boldface marks the column maximum.}}
  {\scriptsize
   \setlength{\tabcolsep}{2.1pt}
   \begin{tabular}{@{}llrrrrrrrr@{}}
   \toprule
   & & \multicolumn{4}{c}{Unconditioned} & \multicolumn{4}{c}{Conditioned} \\
   \cmidrule(lr){3-6}\cmidrule(lr){7-10}
   Release & Dataset & R@1$\downarrow$ & R@5$\downarrow$ & R@10$\downarrow$ & MRR$\downarrow$ & R@1$\downarrow$ & R@5$\downarrow$ & R@10$\downarrow$ & MRR$\downarrow$ \\
   \midrule
   CLIP ViT-B/32 & MIMIC-CXR & $0.00${\tiny$\pm$0.00} & $0.03${\tiny$\pm$0.01} & $0.04${\tiny$\pm$0.01} & $0.04${\tiny$\pm$0.00} & $0.22${\tiny$\pm$0.02} & $0.92${\tiny$\pm$0.05} & $1.90${\tiny$\pm$0.08} & $1.24${\tiny$\pm$0.03} \\
    & CheXpert Plus & $0.00${\tiny$\pm$0.00} & $0.02${\tiny$\pm$0.01} & $0.04${\tiny$\pm$0.01} & $0.04${\tiny$\pm$0.00} & $0.13${\tiny$\pm$0.02} & $0.62${\tiny$\pm$0.05} & $1.18${\tiny$\pm$0.06} & $0.82${\tiny$\pm$0.03} \\
    & ReXGradient-160K & $0.01${\tiny$\pm$0.01} & $0.04${\tiny$\pm$0.01} & $0.09${\tiny$\pm$0.02} & $0.06${\tiny$\pm$0.01} & $0.59${\tiny$\pm$0.05} & $2.46${\tiny$\pm$0.11} & $4.62${\tiny$\pm$0.15} & $2.59${\tiny$\pm$0.07} \\
   \midrule
   CLIP ViT-L/14 & MIMIC-CXR & $0.00${\tiny$\pm$0.00} & $0.04${\tiny$\pm$0.01} & $0.06${\tiny$\pm$0.01} & $0.05${\tiny$\pm$0.00} & $0.25${\tiny$\pm$0.02} & $1.19${\tiny$\pm$0.06} & $2.27${\tiny$\pm$0.09} & $1.41${\tiny$\pm$0.03} \\
    & CheXpert Plus & $0.01${\tiny$\pm$0.01} & $0.03${\tiny$\pm$0.01} & $0.07${\tiny$\pm$0.01} & $0.06${\tiny$\pm$0.01} & $0.18${\tiny$\pm$0.02} & $0.83${\tiny$\pm$0.05} & $1.48${\tiny$\pm$0.07} & $0.98${\tiny$\pm$0.03} \\
    & ReXGradient-160K & $0.04${\tiny$\pm$0.01} & $0.12${\tiny$\pm$0.02} & $0.24${\tiny$\pm$0.03} & $0.14${\tiny$\pm$0.01} & $0.82${\tiny$\pm$0.06} & $3.18${\tiny$\pm$0.12} & $5.63${\tiny$\pm$0.16} & $3.08${\tiny$\pm$0.08} \\
   \midrule
   SigLIP-2 L/16 & MIMIC-CXR & $0.00${\tiny$\pm$0.00} & $0.03${\tiny$\pm$0.01} & $0.06${\tiny$\pm$0.01} & $0.05${\tiny$\pm$0.01} & $0.30${\tiny$\pm$0.03} & $1.30${\tiny$\pm$0.06} & $2.52${\tiny$\pm$0.09} & $1.47${\tiny$\pm$0.04} \\
    & CheXpert Plus & $0.00${\tiny$\pm$0.00} & $0.02${\tiny$\pm$0.01} & $0.04${\tiny$\pm$0.01} & $0.04${\tiny$\pm$0.00} & $0.12${\tiny$\pm$0.02} & $0.59${\tiny$\pm$0.05} & $1.12${\tiny$\pm$0.07} & $0.79${\tiny$\pm$0.03} \\
    & ReXGradient-160K & $0.03${\tiny$\pm$0.01} & $0.10${\tiny$\pm$0.02} & $0.18${\tiny$\pm$0.03} & $0.13${\tiny$\pm$0.01} & $0.83${\tiny$\pm$0.06} & $3.33${\tiny$\pm$0.12} & $5.81${\tiny$\pm$0.16} & $3.14${\tiny$\pm$0.07} \\
   \midrule
   PubMedCLIP & MIMIC-CXR & $0.01${\tiny$\pm$0.00} & $0.03${\tiny$\pm$0.01} & $0.05${\tiny$\pm$0.01} & $0.05${\tiny$\pm$0.01} & $0.24${\tiny$\pm$0.02} & $1.20${\tiny$\pm$0.06} & $2.24${\tiny$\pm$0.09} & $1.37${\tiny$\pm$0.04} \\
    & CheXpert Plus & $0.00${\tiny$\pm$0.00} & $0.03${\tiny$\pm$0.01} & $0.03${\tiny$\pm$0.01} & $0.04${\tiny$\pm$0.00} & $0.11${\tiny$\pm$0.02} & $0.62${\tiny$\pm$0.04} & $1.22${\tiny$\pm$0.06} & $0.84${\tiny$\pm$0.03} \\
    & ReXGradient-160K & $0.03${\tiny$\pm$0.01} & $0.07${\tiny$\pm$0.02} & $0.12${\tiny$\pm$0.02} & $0.10${\tiny$\pm$0.01} & $0.63${\tiny$\pm$0.05} & $2.85${\tiny$\pm$0.12} & $5.20${\tiny$\pm$0.15} & $2.78${\tiny$\pm$0.07} \\
   \midrule
   BiomedCLIP & MIMIC-CXR & $0.04${\tiny$\pm$0.01} & $0.17${\tiny$\pm$0.02} & $0.32${\tiny$\pm$0.03} & $0.19${\tiny$\pm$0.01} & $1.03${\tiny$\pm$0.05} & $3.77${\tiny$\pm$0.11} & $6.50${\tiny$\pm$0.15} & $3.33${\tiny$\pm$0.07} \\
    & CheXpert Plus & $0.04${\tiny$\pm$0.01} & $0.21${\tiny$\pm$0.03} & $0.33${\tiny$\pm$0.03} & $0.21${\tiny$\pm$0.02} & $0.72${\tiny$\pm$0.05} & $2.97${\tiny$\pm$0.11} & $5.15${\tiny$\pm$0.15} & $2.61${\tiny$\pm$0.07} \\
    & ReXGradient-160K & $0.07${\tiny$\pm$0.02} & $0.33${\tiny$\pm$0.04} & $0.62${\tiny$\pm$0.05} & $0.32${\tiny$\pm$0.02} & $1.91${\tiny$\pm$0.09} & $6.69${\tiny$\pm$0.17} & $10.87${\tiny$\pm$0.21} & $5.49${\tiny$\pm$0.10} \\
   \midrule
   MedSigLIP-448 & MIMIC-CXR & $0.29${\tiny$\pm$0.03} & $1.07${\tiny$\pm$0.05} & $1.79${\tiny$\pm$0.07} & $0.95${\tiny$\pm$0.03} & $3.40${\tiny$\pm$0.09} & $11.26${\tiny$\pm$0.19} & $17.59${\tiny$\pm$0.24} & $8.40${\tiny$\pm$0.12} \\
    & CheXpert Plus & $0.25${\tiny$\pm$0.03} & $0.81${\tiny$\pm$0.05} & $1.36${\tiny$\pm$0.07} & $0.77${\tiny$\pm$0.03} & $2.22${\tiny$\pm$0.08} & $7.61${\tiny$\pm$0.17} & $12.42${\tiny$\pm$0.22} & $5.98${\tiny$\pm$0.10} \\
    & ReXGradient-160K & \textbf{$0.40${\tiny$\pm$0.04}} & \textbf{$1.52${\tiny$\pm$0.08}} & \textbf{$2.52${\tiny$\pm$0.10}} & \textbf{$1.28${\tiny$\pm$0.05}} & \textbf{$5.94${\tiny$\pm$0.16}} & \textbf{$16.98${\tiny$\pm$0.27}} & \textbf{$24.69${\tiny$\pm$0.31}} & \textbf{$12.44${\tiny$\pm$0.18}} \\
   \midrule
   BioViL-T & MIMIC-CXR & $0.18${\tiny$\pm$0.02} & $0.68${\tiny$\pm$0.04} & $1.23${\tiny$\pm$0.06} & $0.67${\tiny$\pm$0.03} & $2.87${\tiny$\pm$0.09} & $10.17${\tiny$\pm$0.19} & $16.22${\tiny$\pm$0.26} & $7.65${\tiny$\pm$0.12} \\
    & CheXpert Plus & $0.05${\tiny$\pm$0.01} & $0.21${\tiny$\pm$0.03} & $0.36${\tiny$\pm$0.03} & $0.25${\tiny$\pm$0.02} & $0.92${\tiny$\pm$0.05} & $3.69${\tiny$\pm$0.12} & $6.64${\tiny$\pm$0.18} & $3.28${\tiny$\pm$0.07} \\
    & ReXGradient-160K & $0.08${\tiny$\pm$0.02} & $0.53${\tiny$\pm$0.05} & $0.96${\tiny$\pm$0.06} & $0.49${\tiny$\pm$0.02} & $3.05${\tiny$\pm$0.11} & $10.58${\tiny$\pm$0.22} & $16.54${\tiny$\pm$0.27} & $8.03${\tiny$\pm$0.13} \\
   \midrule
   MONET & MIMIC-CXR & $0.00${\tiny$\pm$0.00} & $0.01${\tiny$\pm$0.01} & $0.03${\tiny$\pm$0.01} & $0.03${\tiny$\pm$0.01} & $0.19${\tiny$\pm$0.02} & $0.93${\tiny$\pm$0.05} & $1.87${\tiny$\pm$0.08} & $1.21${\tiny$\pm$0.03} \\
    & CheXpert Plus & $0.00${\tiny$\pm$0.00} & $0.02${\tiny$\pm$0.01} & $0.05${\tiny$\pm$0.01} & $0.04${\tiny$\pm$0.00} & $0.11${\tiny$\pm$0.02} & $0.55${\tiny$\pm$0.04} & $1.07${\tiny$\pm$0.06} & $0.77${\tiny$\pm$0.03} \\
    & ReXGradient-160K & $0.00${\tiny$\pm$0.00} & $0.01${\tiny$\pm$0.01} & $0.03${\tiny$\pm$0.01} & $0.04${\tiny$\pm$0.00} & $0.52${\tiny$\pm$0.05} & $2.37${\tiny$\pm$0.11} & $4.43${\tiny$\pm$0.14} & $2.48${\tiny$\pm$0.06} \\
   \bottomrule
   \end{tabular}}
\end{table*}

\tableref{tab:fieldmetrics} reports 4 retrieval metrics under both conditions. The 3 highest Recall@1 values belong to MedSigLIP-448, BioViL-T, and BiomedCLIP, in that order, on every dataset and under both conditions. The ordering of the remaining models varies across datasets.

\subsection{Uncertainty in Linkage and Utility}
\label{apd:uncertainty}

\tableref{tab:panelci} lists the 8 releases and the 2 baselines without vision-language encoders, at both operating points. \tableref{tab:variantsci} lists the public checkpoint, the 4 variants, and the matched control, with linkage at $k=1{,}000$ under the primary conjunction.

\begin{table*}[t]
\floatconts
  {tab:panelci}
  {\caption{Precision at $k=1{,}000$: bootstrap mean $\pm$ standard deviation [95\% percentile interval], percent, $\downarrow$. Shading marks zero-width intervals.}}
  {\scriptsize
   \setlength{\tabcolsep}{3.2pt}
   \begin{tabular}{@{}llll@{}}
   \toprule
   Release & Dataset & Unconditioned$\downarrow$ & Conditioned$\downarrow$ \\
   \midrule
   CLIP ViT-B/32 & MIMIC-CXR & \cellcolor{gray!14}0.00 $\pm$ 0.00 [0.00, 0.00] & 0.20 $\pm$ 0.15 [0.00, 0.50] \\
    & CheXpert Plus & \cellcolor{gray!14}0.00 $\pm$ 0.00 [0.00, 0.00] & 0.00 $\pm$ 0.00 [0.00, 0.00] \\
    & ReXGradient-160K & \cellcolor{gray!14}0.00 $\pm$ 0.00 [0.00, 0.00] & 1.90 $\pm$ 0.50 [1.00, 2.90] \\
   \midrule
   CLIP ViT-L/14 & MIMIC-CXR & \cellcolor{gray!14}0.00 $\pm$ 0.00 [0.00, 0.00] & 0.40 $\pm$ 0.20 [0.10, 0.80] \\
    & CheXpert Plus & \cellcolor{gray!14}0.00 $\pm$ 0.00 [0.00, 0.00] & 0.41 $\pm$ 0.21 [0.10, 0.90] \\
    & ReXGradient-160K & 0.11 $\pm$ 0.11 [0.00, 0.30] & 3.63 $\pm$ 0.73 [2.30, 5.10] \\
   \midrule
   SigLIP-2 L/16 & MIMIC-CXR & \cellcolor{gray!14}0.00 $\pm$ 0.00 [0.00, 0.00] & 0.87 $\pm$ 0.35 [0.30, 1.60] \\
    & CheXpert Plus & \cellcolor{gray!14}0.00 $\pm$ 0.00 [0.00, 0.00] & 0.10 $\pm$ 0.10 [0.00, 0.30] \\
    & ReXGradient-160K & 0.10 $\pm$ 0.10 [0.00, 0.30] & 3.01 $\pm$ 0.59 [1.90, 4.20] \\
   \midrule
   PubMedCLIP & MIMIC-CXR & \cellcolor{gray!14}0.00 $\pm$ 0.00 [0.00, 0.00] & 0.81 $\pm$ 0.29 [0.30, 1.40] \\
    & CheXpert Plus & \cellcolor{gray!14}0.00 $\pm$ 0.00 [0.00, 0.00] & 0.19 $\pm$ 0.14 [0.00, 0.50] \\
    & ReXGradient-160K & 0.10 $\pm$ 0.10 [0.00, 0.30] & 3.11 $\pm$ 0.58 [2.00, 4.30] \\
   \midrule
   BiomedCLIP & MIMIC-CXR & \cellcolor{gray!14}0.00 $\pm$ 0.00 [0.00, 0.00] & 3.74 $\pm$ 0.64 [2.60, 5.10] \\
    & CheXpert Plus & \cellcolor{gray!14}0.00 $\pm$ 0.00 [0.00, 0.00] & 1.31 $\pm$ 0.35 [0.60, 2.10] \\
    & ReXGradient-160K & 0.10 $\pm$ 0.10 [0.00, 0.30] & 9.13 $\pm$ 1.05 [7.30, 11.10] \\
   \midrule
   MedSigLIP-448 & MIMIC-CXR & 0.85 $\pm$ 0.29 [0.30, 1.40] & 14.94 $\pm$ 1.18 [12.50, 17.30] \\
    & CheXpert Plus & 0.36 $\pm$ 0.19 [0.10, 0.70] & 6.42 $\pm$ 0.77 [5.00, 7.90] \\
    & ReXGradient-160K & 0.90 $\pm$ 0.29 [0.40, 1.50] & 28.74 $\pm$ 1.57 [25.80, 31.90] \\
   \midrule
   BioViL-T & MIMIC-CXR & 0.20 $\pm$ 0.14 [0.00, 0.50] & 8.96 $\pm$ 0.87 [7.30, 10.70] \\
    & CheXpert Plus & \cellcolor{gray!14}0.00 $\pm$ 0.00 [0.00, 0.00] & 1.40 $\pm$ 0.37 [0.70, 2.10] \\
    & ReXGradient-160K & 0.10 $\pm$ 0.10 [0.00, 0.30] & 10.37 $\pm$ 1.06 [8.50, 12.60] \\
   \midrule
   MONET & MIMIC-CXR & \cellcolor{gray!14}0.00 $\pm$ 0.00 [0.00, 0.00] & 0.38 $\pm$ 0.20 [0.10, 0.80] \\
    & CheXpert Plus & \cellcolor{gray!14}0.00 $\pm$ 0.00 [0.00, 0.00] & 0.26 $\pm$ 0.17 [0.00, 0.60] \\
    & ReXGradient-160K & \cellcolor{gray!14}0.00 $\pm$ 0.00 [0.00, 0.00] & 1.53 $\pm$ 0.38 [0.80, 2.30] \\
   \midrule
   14-label classifier & MIMIC-CXR & \cellcolor{gray!14}0.00 $\pm$ 0.00 [0.00, 0.00] & 0.41 $\pm$ 0.20 [0.10, 0.80] \\
    & CheXpert Plus & 0.10 $\pm$ 0.10 [0.00, 0.30] & 0.38 $\pm$ 0.21 [0.10, 0.80] \\
    & ReXGradient-160K & \cellcolor{gray!14}0.00 $\pm$ 0.00 [0.00, 0.00] & 1.91 $\pm$ 0.46 [1.10, 2.90] \\
   \midrule
   Published attributes & MIMIC-CXR & 1.11 $\pm$ 0.33 [0.50, 1.80] & N/A \\
    & CheXpert Plus & 0.30 $\pm$ 0.17 [0.00, 0.70] & N/A \\
    & ReXGradient-160K & 4.37 $\pm$ 0.67 [3.10, 5.70] & N/A \\
   \bottomrule
   \end{tabular}}
\end{table*}
\begin{table*}[t]
\floatconts
  {tab:variantsci}
  {\caption{Head variants at $k=1{,}000$: bootstrap mean $\pm$ standard deviation [95\% percentile interval], in percent. The public checkpoint row is shaded.}}
  {\scriptsize
   \setlength{\tabcolsep}{3.2pt}
   \begin{tabular}{@{}llll@{}}
   \toprule
   Variant & Dataset & Conditioned linkage$\downarrow$ & Zero-shot AUROC (MIMIC-CXR)$\uparrow$ \\
   \midrule
   \rowcolor{gray!12} Public checkpoint & MIMIC-CXR & 8.96 $\pm$ 0.87 [7.30, 10.70] & 68.86 $\pm$ 2.53 [64.27, 73.84] \\
    & CheXpert Plus & 1.40 $\pm$ 0.37 [0.70, 2.10] &  \\
    & ReXGradient-160K & 10.37 $\pm$ 1.06 [8.50, 12.60] &  \\
   \midrule
   Matched non-private control & MIMIC-CXR & 28.97 $\pm$ 1.53 [25.90, 31.90] & 69.66 $\pm$ 2.75 [64.35, 74.96] \\
    & CheXpert Plus & 10.76 $\pm$ 0.99 [8.90, 12.70] &  \\
    & ReXGradient-160K & 27.00 $\pm$ 1.50 [24.10, 29.90] &  \\
   \midrule
   DP heads & MIMIC-CXR & 18.26 $\pm$ 1.30 [15.70, 20.70] & 69.89 $\pm$ 2.72 [64.51, 75.12] \\
    & CheXpert Plus & 5.82 $\pm$ 0.76 [4.40, 7.40] &  \\
    & ReXGradient-160K & 18.74 $\pm$ 1.29 [16.30, 21.20] &  \\
   \midrule
   DP heads, in-batch loss & MIMIC-CXR & 20.05 $\pm$ 1.33 [17.50, 22.70] & 69.12 $\pm$ 2.29 [64.97, 73.60] \\
    & CheXpert Plus & 3.63 $\pm$ 0.58 [2.50, 4.80] &  \\
    & ReXGradient-160K & 18.66 $\pm$ 1.35 [16.10, 21.50] &  \\
   \midrule
   Bypass heads on public data & MIMIC-CXR & 45.55 $\pm$ 1.72 [42.30, 48.80] & 69.70 $\pm$ 2.65 [64.91, 75.10] \\
    & CheXpert Plus & 18.11 $\pm$ 1.27 [15.60, 20.70] &  \\
    & ReXGradient-160K & 31.35 $\pm$ 1.67 [28.40, 34.70] &  \\
   \midrule
   Random orthogonal remapping & MIMIC-CXR & 0.16 $\pm$ 0.13 [0.00, 0.50] & 48.38 $\pm$ 2.48 [43.54, 53.19] \\
    & CheXpert Plus & 0.10 $\pm$ 0.10 [0.00, 0.30] &  \\
    & ReXGradient-160K & 1.20 $\pm$ 0.56 [0.30, 2.40] &  \\
   \bottomrule
   \end{tabular}}
\end{table*}

We resample patients because image-level resampling would treat correlated observations as independent and could understate uncertainty.

Of the 30 unconditioned intervals in \tableref{tab:panelci}, 17 have zero width. These cases have no observed recoveries among selected queries. Rare positive bootstrap replicates can still produce a small positive mean while both central percentile endpoints remain zero. Such an interval does not establish zero population risk. The paired defense tests are computable at both operating points. Under conditioning, MedSigLIP-448 has nonoverlapping intervals with every other panel model on all 3 datasets. BioViL-T and BiomedCLIP have nonoverlapping intervals only on MIMIC-CXR. Among CLIP ViT-B/32, CLIP ViT-L/14, SigLIP-2, PubMedCLIP, and MONET, 1 of 30 comparisons has nonoverlapping intervals. Interval overlap is descriptive and does not replace a test of the difference.

\subsection{Every Conditioning Level}
\label{apd:grid}

\paragraph{Deriving Each Attribute.} Sex uses the supplied field, with a separate category for records that are neither male nor female. Age is grouped into 5-year bins. MIMIC-CXR and ReXGradient-160K include fractional ages, while CheXpert Plus supplies integer ages. Projection uses the supplied view, restricted to PA and AP on MIMIC-CXR and ReXGradient-160K and to the single Frontal value on CheXpert Plus. Study year is extracted from the supplied date. Site uses ReXGradient-160K's institution field. We derive every attribute from the pool manifest with the same mapping for queries and candidate reports.

\paragraph{CheXpert Plus: Identical Primary and Tightest Conjunctions.} CheXpert Plus publishes no date and has a single projection category. Therefore, the primary and tightest conjunctions use the same sex and age-band filters.

\begin{figure}[t]
\floatconts
  {fig:doseresponse}
  {\caption{Pool size and precision. The diagonal is $100/N_{\mathrm{med}}$.}}
  {\includegraphics[width=3.18in]{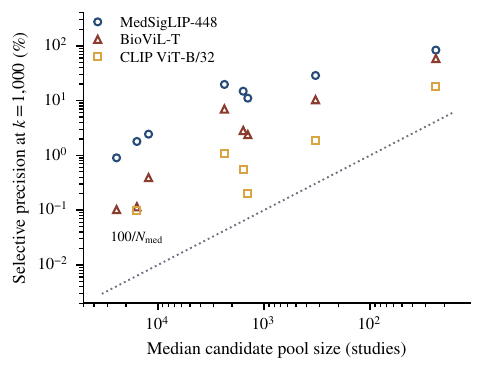}}
\end{figure}

\begin{table*}[t]
\floatconts
  {tab:grid}
  {\caption{Every conditioning level, the resulting pool size, and the selective precision at $k=1{,}000$ in percent, with the unconditioned row shaded. Singletons are queries left with 1 candidate.}}
  {\scriptsize
   \setlength{\tabcolsep}{4pt}
   \begin{tabular}{@{}llrrrrr@{}}
   \toprule
   Dataset & Attributes given & Pool & Singletons & BioViL-T$\downarrow$ & MedSigLIP-448$\downarrow$ & CLIP ViT-B/32$\downarrow$ \\
   \midrule
   \rowcolor{gray!12} MIMIC-CXR & Nothing & 38{,}686 & N/A & 0.20 & 0.85 & 0.00 \\
    & Sex & 18{,}799 & 0 & 0.89 & 2.56 & 0.00 \\
    & Age band & 3{,}013 & 1 & 3.94 & 8.99 & 0.10 \\
    & Projection & 22{,}596 & 0 & 0.40 & 1.32 & 0.00 \\
    & Study year & 466 & 0 & 13.23 & 20.95 & 0.77 \\
    & Primary conjunction & 862 & 1 & 8.96 & 14.94 & 0.20 \\
    & Tightest conjunction & 12 & 892 & 94.70 & 87.35 & 30.78 \\
    & Pathology-matched pool & 10{,}000 & 0 & 0.30 & 2.39 & 0.00 \\
   \midrule
   \rowcolor{gray!12} CheXpert Plus & Nothing & 31{,}631 & N/A & 0.00 & 0.36 & 0.00 \\
    & Sex & 18{,}458 & 1 & 0.10 & 0.80 & 0.00 \\
    & Age band & 2{,}723 & 0 & 0.80 & 4.99 & 0.00 \\
    & Projection & 31{,}631 & 0 & 0.00 & 0.36 & 0.00 \\
    & Study year & 31{,}631 & 0 & 0.00 & 0.36 & 0.00 \\
    & Primary conjunction & 1{,}203 & 1 & 1.40 & 6.42 & 0.00 \\
    & Tightest conjunction & 1{,}203 & 1 & 1.40 & 6.42 & 0.00 \\
    & Pathology-matched pool & 10{,}000 & 0 & 0.10 & 0.61 & 0.00 \\
   \midrule
   \rowcolor{gray!12} ReXGradient-160K & Nothing & 24{,}587 & N/A & 0.10 & 0.90 & 0.00 \\
    & Sex & 12{,}274 & 0 & 0.39 & 2.44 & 0.00 \\
    & Age band & 1{,}423 & 0 & 2.40 & 11.08 & 0.20 \\
    & Projection & 15{,}808 & 0 & 0.12 & 1.79 & 0.10 \\
    & Study year & 1{,}572 & 0 & 2.84 & 14.83 & 0.55 \\
    & Site & 2{,}364 & 3 & 7.02 & 19.73 & 1.10 \\
    & Primary conjunction & 327 & 4 & 10.37 & 28.74 & 1.90 \\
    & Tightest conjunction & 24 & 190 & 58.88 & 83.63 & 18.18 \\
    & Pathology-matched pool & 10{,}000 & 0 & 0.10 & 1.48 & 0.10 \\
   \bottomrule
   \end{tabular}}
\end{table*}

\tableref{tab:grid} covers the 3 releases evaluated on the full grid. \figureref{fig:doseresponse} plots selective precision and median pool size for each conditioning level on ReXGradient-160K. Its diagonal is the reciprocal of the median pool size, $N_{\mathrm{med}}$, in percent. This reference differs from the selected-query guessing expectation. Pool reduction depends on the dataset's metadata distribution. Sex reduces the median pool by about 50\% on all 3 datasets. An age band reduces it by about an order of magnitude. Projection does not change the pool on CheXpert Plus. On ReXGradient-160K, age band produces the largest single-attribute reduction, from 24{,}587 to 1{,}423 studies, followed by calendar year at 1{,}572. The pathology-matched pool of \appendixref{apd:instrument} contains 10{,}000 candidates, which is smaller than the unconditioned pool and larger than any attribute conjunction.

\subsection{The Content of a Recovered Report}
\label{apd:exposure}

\begin{figure}[t]
\floatconts
  {fig:exposurebars}
  {\caption{Recovered reports, ReXGradient-160K. Bars are descriptive differences.}}
  {\includegraphics[width=3.18in]{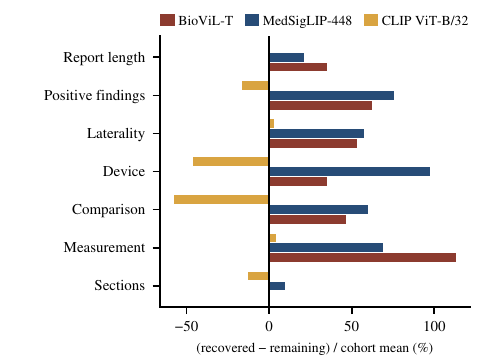}}
\end{figure}

\paragraph{The Contrast.} We compare covariate means for correctly matched queries with those for the remaining queries in the same conditioned dataset. Each query image is an observation. Therefore, a report can appear more than once. Count differences retain their original units, and binary differences are percentage points. These contrasts are descriptive. We report no significance tests because queries from the same patient are dependent. At the primary conjunction, this analysis evaluates BioViL-T, MedSigLIP-448, and CLIP ViT-B/32. The last model is a near-chance comparison, without establishing a causal effect of conditioning. \tableref{tab:exposure} contains all 8 measured covariates.

\begin{table*}[t]
\floatconts
  {tab:exposure}
  {\caption{Descriptive differences between recovered and remaining queries. Counts retain their units, and binary differences are percentage points.}}
  {\scriptsize
   \setlength{\tabcolsep}{3pt}
   \begin{tabular}{@{}lrrrrrrrrr@{}}
   \toprule
   & \multicolumn{3}{c}{MIMIC-CXR} & \multicolumn{3}{c}{CheXpert Plus} & \multicolumn{3}{c}{ReXGradient-160K} \\
   \cmidrule(lr){2-4}\cmidrule(lr){5-7}\cmidrule(lr){8-10}
   Report covariate & BioViL-T & \shortstack[r]{MedSigLIP-\\448} & \shortstack[r]{CLIP\\ViT-B/32} & BioViL-T & \shortstack[r]{MedSigLIP-\\448} & \shortstack[r]{CLIP\\ViT-B/32} & BioViL-T & \shortstack[r]{MedSigLIP-\\448} & \shortstack[r]{CLIP\\ViT-B/32} \\
   \midrule
Report length in tokens & $+$15.85 & $+$2.52 & $-$1.39 & $+$6.85 & $-$5.99 & $+$3.99 & $+$18.52 & $+$11.31 & $-$0.35 \\
Section count & $+$0.16 & $-$0.10 & $-$0.07 & $+$0.10 & $-$0.00 & $+$0.01 & $-$0.01 & $+$0.13 & $-$0.16 \\
Positive findings & $-$0.04 & $+$0.21 & $-$0.36 & $-$0.62 & $-$0.03 & $+$0.10 & $+$0.98 & $+$1.19 & $-$0.25 \\
States laterality & $+$13.96 & $+$12.09 & $+$0.51 & $-$2.79 & $-$0.66 & $-$0.53 & $+$27.35 & $+$29.39 & $+$1.57 \\
Mentions a device & $-$9.16 & $+$3.15 & $-$9.50 & $-$24.56 & $-$7.83 & $-$0.16 & $+$6.43 & $+$17.75 & $-$8.37 \\
Compares with a prior & $+$2.03 & $+$4.02 & $-$7.18 & $-$0.75 & $+$0.75 & $+$5.69 & $+$7.26 & $+$9.30 & $-$8.97 \\
Contains a measurement & $-$0.14 & $-$2.42 & $+$0.23 & $+$1.29 & $-$2.27 & $-$3.06 & $+$6.47 & $+$3.94 & $+$0.24 \\
\shortstack[l]{De-identification\\placeholders} & $+$0.35 & $+$0.02 & $-$0.09 & $-$0.01 & $-$0.01 & $+$0.05 & $-$0.00 & $-$0.00 & $-$0.00 \\
   \bottomrule
   \end{tabular}}
\end{table*}

\figureref{fig:exposurebars} compares 7 of the 8 covariates on ReXGradient-160K. Each difference is divided by the mean of its entire conditioned query set. The placeholder count is omitted because only 1 query contains a detected run. The sign agreement of the separate patient-clustered margin regressions is reported in \appendixref{apd:map}.

\section{Robustness of Query Selection and Model Comparisons}
\label{apd:further}

\subsection{The Decision Statistic}
\label{apd:statistic2}

\begin{table}[t]
\floatconts
  {tab:statistic}
  {\caption{Decision statistics at $k=1{,}000$, percent. Boldface marks the row maximum.}}
  {\scriptsize
   \setlength{\tabcolsep}{2.8pt}
   \begin{tabular}{@{}llrrrr@{}}
   \toprule
   Release & Dataset & Margin$\downarrow$ & Top-1$\downarrow$ & Ratio$\downarrow$ & Undefined \\
   \midrule
   MedSigLIP-448 & MIMIC & \textbf{14.94} & 8.80 & 14.72 & 0.01 \\
    & CheXpert & \textbf{6.42} & 6.02 & 6.14 & 0.01 \\
    & ReXGrad. & \textbf{28.74} & 17.44 & 27.93 & 0.02 \\
   \midrule
   BioViL-T & MIMIC & \textbf{8.96} & 7.12 & 8.72 & 0.00 \\
    & CheXpert & 1.40 & \textbf{1.66} & 1.38 & 0.00 \\
    & ReXGrad. & \textbf{10.37} & 3.28 & 10.30 & 0.00 \\
   \midrule
   BiomedCLIP & MIMIC & 3.74 & 1.37 & \textbf{3.88} & 0.00 \\
    & CheXpert & 1.31 & 0.95 & \textbf{1.52} & 0.00 \\
    & ReXGrad. & 9.13 & 2.22 & \textbf{9.65} & 0.00 \\
   \bottomrule
   \end{tabular}}
\end{table}

We compare 3 statistics that an adversary can compute from the similarity scores alone. Beside the margin of \equationref{eq:margin}, the top-1 score is the best similarity alone. The ratio is the best divided by the second-best and is evaluated only when the second-best similarity exceeds $10^{-12}$.

\tableref{tab:statistic} compares all 3 statistics at the primary conjunction. The top-1 score depends on the query-specific similarity scale as well as on the separation of the best match, whereas the margin subtracts the second-best score. For the 3 releases in \tableref{tab:statistic}, the undefined share of the ratio is at most 0.02\% at the untransformed report. This fraction is as high as 31\% under the label-only template, which replaces the narrative with fixed label sentences. We use the margin for all primary results.

\subsection{Ties and Uniquely Identified Queries}
\label{apd:ties}

A singleton query is left out of the ranked analysis because a margin needs a second candidate. \tableref{tab:grid} reports those counts per conditioning level. The sensitivity analysis flags a query when the selected top report belongs to another study and the true report's score is within $10^{-6}$ of the maximum. This rule does not remove ties for which the selected report is the true report. It also flags near ties within that tolerance. Counts are at most 1 for the untransformed report. Under the label-only template on MIMIC-CXR, the counts are 1{,}964 for CLIP ViT-B/32 and 1{,}807 for MedSigLIP-448. Where at least 10 queries remain after exclusion, we recompute Recall@1. \sectionref{sec:defense} quotes these sensitivity estimates.

\subsection{Several Images of One Study}
\label{apd:studylevel}

A study with several images of the same projection permits a combined attack on its report. We group images by study and projection so that every image in a group has the same candidate set. We average their similarity scores for each candidate and rank groups by the resulting margin. The comparator uses the image with the largest individual margin in each group. The 2 attacks select from the same groups. Pooling across a patient's different studies has no single target report. \tableref{tab:studylevel} reports the comparison at the primary conjunction and a budget of 100 groups.

\begin{table*}[t]
\floatconts
  {tab:studylevel}
  {\caption{Pooled and single-image precision (percent), primary conjunction, $k=100$. The difference is in percentage points.}}
  {\scriptsize
   \setlength{\tabcolsep}{4pt}
   \begin{tabular}{@{}llrrrrr@{}}
   \toprule
   Release & Dataset & Groups & Pooled & Single & Difference & $p$ \\
   \midrule
   MedSigLIP-448 & MIMIC-CXR & 4239 & 18.24 & 20.63 & $-$2.39 & $>$0.999 \\
    & CheXpert Plus & 572 & 3.02 & 6.47 & $-$3.45 & $>$0.999 \\
    & ReXGradient-160K & 872 & 16.70 & 13.04 & $+$3.67 & 0.524 \\
   \midrule
   BioViL-T & MIMIC-CXR & 4239 & 6.78 & 8.34 & $-$1.56 & $>$0.999 \\
    & CheXpert Plus & 572 & 1.89 & 2.42 & $-$0.53 & $>$0.999 \\
    & ReXGradient-160K & 872 & 9.13 & 7.15 & $+$1.98 & 0.624 \\
   \midrule
   CLIP ViT-B/32 & MIMIC-CXR & 4239 & 0.00 & 0.00 & $+$0.00 & $>$0.999 \\
    & CheXpert Plus & 572 & 0.00 & 0.00 & $+$0.00 & $>$0.999 \\
    & ReXGradient-160K & 872 & 3.06 & 5.38 & $-$2.31 & 0.524 \\
   \bottomrule
   \end{tabular}}
\end{table*}

\subsection{Agreement of the Release Rankings}
\label{apd:tau}

\begin{table*}[t]
\floatconts
  {tab:tau}
  {\caption{Kendall's $\tau$ and adjusted $p$ for rankings of 12 releases. Red marks $\tau < 0.5$.}}
  {\scriptsize
   \setlength{\tabcolsep}{3pt}
   \begin{tabular}{@{}llrrrrrr@{}}
   \toprule
   Condition & Dataset & $\tau$, $k=10$ & $p$ & $\tau$, $k=100$ & $p$ & $\tau$, $k=1{,}000$ & $p$ \\
   \midrule
   Primary conjunction & MIMIC-CXR & 0.790 & $<$0.001 & 0.931 & $<$0.001 & 0.939 & $<$0.001 \\
    & CheXpert Plus & \cellcolor{red!16}0.445 & 0.050 & 0.574 & 0.016 & 0.697 & 0.003 \\
    & ReXGradient-160K & 0.697 & $<$0.001 & 0.788 & $<$0.001 & 0.909 & $<$0.001 \\
   \midrule
   Unconditioned & MIMIC-CXR & \cellcolor{red!16}0.498 & 0.045 & 0.721 & 0.004 & 0.758 & 0.004 \\
    & CheXpert Plus & \cellcolor{red!16}0.460 & 0.103 & \cellcolor{red!16}0.406 & 0.109 & 0.684 & 0.015 \\
    & ReXGradient-160K & \cellcolor{red!16}0.244 & 0.336 & 0.536 & 0.038 & 0.859 & $<$0.001 \\
   \midrule
   \multicolumn{8}{@{}l}{\textbf{Inverted pairs among 66, primary conjunction}} \\
   & MIMIC-CXR & 4 & & 2 & & 2 & \\
   & CheXpert Plus & 16 & & 13 & & 10 & \\
   & ReXGradient-160K & 10 & & 7 & & 3 & \\
   \bottomrule
   \end{tabular}}
\end{table*}

\tableref{tab:tau} compares the Recall@1 and selective precision rankings of the 8 panel models and 4 release variants. Rankings use full-precision estimates, with the matched non-private control excluded. We report Kendall's $\tau$, its adjusted $p$, and inverted pairs among the 66 possible model pairs.

At the primary conjunction, rank agreement is positive at every budget but varies by dataset. Smaller budgets can produce more ties and inversions because selective precision uses fewer queries. The $p$ values concern rank association across the 12 model-level estimates. They do not propagate uncertainty from the shared evaluation patients or establish agreement beyond these releases.

\subsection{Cross-Model Overlap}
\label{apd:overlap}

\begin{figure}[t]
\floatconts
  {fig:overlapstrip}
  {\caption{Release pairs, primary conjunction. Clinical pairs have Recall@1 folds $\geq2$.}}
  {\includegraphics[width=3.18in]{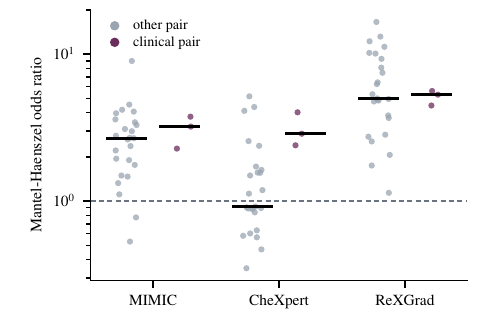}}
\end{figure}

We also describe overlap in correctly matched query images across models. Candidate-pool size can confound this association, and repeated images from the same patient are dependent. The overlap estimates are descriptive because the test treats queries as independent.

Without conditioning, 63 of the 84 release pairs have recoveries for each model, permitting an odds-ratio estimate. The largest estimate is 3.49. Under the primary conjunction, all 84 pairs permit estimation. The 9 pairs drawn from the 3 clinical models have estimates from 2.28 to 5.62 and share 7 to 204 correctly matched queries.

We stratify by candidate-pool size in up to 10 quantile bands and compute a Mantel-Haenszel odds ratio \citep{Mantel1959}, adding 0.5 to its pooled numerator and denominator for stability. This adjustment does not remove variation in difficulty within a band or dependence among a patient's queries. We interpret the estimates descriptively. In \figureref{fig:overlapstrip}, the clinical-pair medians lie within the range of the other pairs on each dataset. The separate patient-clustered regressions in \appendixref{apd:map} provide the report-covariate analysis.

\section{Limits of the Access, Data, and Defense Assumptions}
\label{apd:limits}

The size and composition of an attacker's report collection affect both candidate ranking and query selection. At a budget of 100, enlarging the collection to all studies in a dataset lowers precision from 67.68\% to 37.23\% on ReXGradient-160K and from 35.38\% to 14.38\% on MIMIC-CXR (\sectionref{sec:costs}). This empirical reduction is not a general upper bound for health-system collections because adding candidates can also change which queries are selected. Additional attributes, such as referral date or department, could produce smaller pools than those evaluated here. A hospital needs a measurement on its collection and available attributes.

MIMIC-CXR's shifted study year contains patient-specific date-shift information and cannot be interpreted as acquisition calendar year. Its year-conditioned result is a diagnostic analysis, excluded from realistic attack-rate claims. ReXGradient-160K publishes calendar year and supplies the strongest evaluated metadata setting. We evaluate the full conditioning grid for 3 models and compare all 8 panel models at the primary conjunction.

We ran no reader study and cannot estimate the additional linkage risk relative to a trained reader. The attribute and classifier baselines in \appendixref{apd:nomodel} are computational comparisons. Their performance does not bound human performance, and the classifier covers only 12 of the 14 report findings. Text-to-image utility retrieval uses 5{,}000 distinct studies. Linkage evaluation uses all eligible queries before budget selection.

The privacy discussion concerns head finetuning conditional on fixed encoders and validation negatives (\appendixref{apd:dp}). It provides no guarantee for encoder pretraining or the validation bank. All defense variants use BioViL-T, limiting generalization to other architectures. Different checkpoint-selection criteria and batch sampling confound attribution of private-versus-control differences to clipping and noise.

The pretraining-overlap limitation is discussed in \sectionref{sec:conclusion}. The dataset split controls which records enter head finetuning, without establishing a patient-disjoint evaluation of the pretrained encoders.

The content axes in \appendixref{apd:transforms} detect mentions using fixed text patterns. They can remain positive after a change in laterality or a measurement value. The report might then contradict the original despite retaining a mention on that axis. \sectionref{sec:defense} describes the labeling limitations and the absence of clinician validation.

All results concern English-language chest radiography and use 3 datasets that publish the image-report pairing. Because these datasets reveal the pairing, we can measure whether a candidate linkage is correct. They support the evaluation and do not represent the privacy scenario itself, in which the reports are withheld. We do not assume that these results generalize to another imaging modality, another language, or a record type whose free text is structured differently. We do not assess the removal of direct patient identifiers from reports.

\section{Practical Implications for a Releasing Hospital}
\label{apd:practice}

A hospital can count candidate reports for each proposed combination of released attributes before running a model. When the true report is in the collection and its metadata are correct, a singleton set identifies the report through attributes alone. We count these queries separately because a margin requires at least 2 candidates. With missing reports or erroneous attributes, a singleton candidate does not necessarily identify the correct report.

The evaluation requires images, a report collection, their metadata, and true pairings for scoring. A hospital can embed each item once and reuse the similarities across budgets and attribute settings. Reporting both Recall@1 and selective linkage precision distinguishes whole-pool retrieval from a selective attack. The candidate collection should approximate the attacker's access. The tested whole-dataset collections give lower precision than the test-split collections (\sectionref{sec:costs}).

Text-pattern counts indicate how often a report mentions an attribute, although a mention does not verify its value. The uniform-guess expectation must use the selected queries. Removing correct reports from part of the collection tests precision under incomplete access and records how often selected queries have no possible correct match.

The tested random maps lower cross-modal linkage and zero-shot utility while preserving within-image geometry. The label template has linkage intervals below those of the original reports but removes details outside the 14-finding label space (\sectionref{sec:defense}). Choosing a clinically acceptable release requires utility and clinician assessments beyond these measurements.

This study measures linkage risk and does not evaluate clinical safety or clinical benefit.

\end{document}